\documentclass{article}

\usepackage[preprint]{neurips_2026}  

\usepackage[utf8]{inputenc}
\usepackage[T1]{fontenc}
\usepackage{microtype}

\usepackage{amsfonts}
\usepackage{amsmath}
\usepackage{amssymb}
\usepackage{nicefrac}

\usepackage{booktabs}
\usepackage{multirow}
\usepackage{makecell}
\usepackage{siunitx}
\usepackage{wrapfig}
\usepackage{graphicx}
\usepackage{enumitem}
\usepackage{xcolor}
\usepackage{placeins}
\usepackage{adjustbox}
\usepackage{pifont}

\usepackage{hyperref}
\usepackage{url}
\hypersetup{
    colorlinks=true,
    linkcolor=blue!50!black,
    citecolor=blue!50!black,
    urlcolor=blue!50!black,
    breaklinks=true
}

\usepackage{tikz}
\usetikzlibrary{positioning, arrows.meta, fit, calc, decorations.pathreplacing}
\usepackage{svg}

\newcommand{\modelname}{Chronicle}

\definecolor{cblue}{HTML}{4A7FBF}
\definecolor{cbluelt}{HTML}{D6E4F0}
\definecolor{cteal}{HTML}{2E8B7A}
\definecolor{cteallt}{HTML}{D0ECE6}
\definecolor{cpurple}{HTML}{6B5DA8}
\definecolor{cpurplelt}{HTML}{E0DBEF}
\definecolor{ccoral}{HTML}{C06040}
\definecolor{ccorallt}{HTML}{F2DDD4}
\definecolor{cgray}{HTML}{6B6B6B}
\definecolor{cgraylt}{HTML}{E8E8E8}

\title{\textbf{\modelname{}: A Multimodal Foundation Model\\for Joint Language and Time Series Understanding}}

\author{%
  Paul Quinlan$^{1,2}$\thanks{Correspondence to: \texttt{paul.quinlan@inertialai.com}} \quad Jeremy Levasseur$^{1}$ \quad Qingguo Li$^{3}$ \quad Xiaodan Zhu$^{2}$ \\[0.5em]
  $^{1}$InertialAI \\
  $^{2}$Department of Electrical and Computer Engineering, Queen's University \\
  $^{3}$Department of Mechanical and Materials Engineering, Queen's University \\
}

\date{}

\begin{document}

\maketitle

\begin{abstract}
Real-world time series come with text: metadata, descriptions, news,
reports. Yet time series foundation models process numerical sequences
in isolation, and the multimodal text-and-time-series models that
attempt to bridge the two all adapt a pretrained language model post
hoc, inheriting representations shaped without ever seeing temporal
data. These models are also evaluated almost exclusively against other
multimodal baselines, not against the strongest unimodal foundation
models in either domain, leaving open whether joint training is needed
at all.
We present \textbf{\modelname{}}, a compact 324M-parameter decoder-only
transformer trained from scratch on natural language and time series
within a single unified architecture. Both modalities share the same
transformer blocks, attention mechanism, and residual stream; the bulk
of pretraining uses unimodal batches so cross-modal capability emerges
purely from shared parameters, with a short alignment stage that
interleaves the two. To our knowledge, \modelname{} is the first model
jointly pretrained on text and time series from scratch, and the first
multimodal model evaluated against dedicated foundation models in both
domains. It matches Gemma-3-270M-PT on 19 NLU tasks, sets a new bar
for frozen-embedding time series classification on 24 UCR/UEA datasets,
and produces multimodal forecasts on Time-MMD that beat every
supervised fusion baseline, all from a single backbone.
\end{abstract}

\section{Introduction}
\label{sec:intro}
 
Time series foundation models (TSFMs) have transformed forecasting into an inference-only pipeline: a single pretrained model can be applied zero-shot across diverse domains~\citep{ansari2024chronos, das2024timesfm, woo2024moirai, wen2026patchtst-fm}.
Models such as Chronos-2~\citep{ansari2025chronos2} and PatchTST-FM~\citep{wen2026patchtst-fm} set a high bar on standardized benchmarks like GIFT-Eval~\citep{jain2024gifteval}.
Yet these models remain narrow specialists that process numerical sequences in isolation, with no mechanism to incorporate the textual context (metadata, domain knowledge, anomaly descriptions) that accompanies virtually every real-world time series.
 
A growing body of work has attempted to bridge this gap by connecting language models to time series, but these efforts share three systematic limitations.
First, every existing approach starts from a pretrained language model and adapts it post hoc.
LLMTIME~\citep{gruver2023llmtime} and GPT4MTS~\citep{jia2024gpt4mts} query frozen LLMs directly; Time-LLM~\citep{jin2024timellm} and GPT4TS~\citep{zhou2023gpt4ts} add lightweight adapters; ChatTS~\citep{xie2024chatts}, ChatTime~\citep{wang2024chattime}, and MSE-ITT~\citep{koval2025msitt} fine-tune large backbones (Qwen2.5-14B, LLaMA-2, LLaMA-3-8B); and MoAT~\citep{lee2024moat} and TaTs~\citep{li2025tats} fuse separate encoders via learned heads (Table~\ref{tab:positioning}).
Because these models were pretrained on text alone, their internal representations were shaped without any exposure to temporal data, and the time series modality must adapt to a representational space not designed for it.
No prior work has trained a single model from scratch on both modalities, allowing text and time series to shape each other's representations from the beginning of training.
 
Second, these models are evaluated almost exclusively against other multimodal or task-specific baselines, not against state-of-the-art unimodal foundation models in either domain.
Recent surveys~\citep{ liu2025crossmodalitymodelingtimeseries} have noted this gap, and \citet{zhang2025whenmm} found that the benefits of multimodality are ``highly condition-dependent,'' underscoring the need for rigorous unimodal baselines.
Third, many of these models, particularly ChatTS, ChatTime, Chat-TS~\citep{quinlan2026chattsenhancingmultimodalreasoning}, and MSE-ITT, target conversational reasoning about time series (question answering, summarization, explanation) and require large backbones (8 to 14B parameters) to support instruction following.
This conflates two distinct goals: building general-purpose temporal \emph{representations} versus building temporal \emph{reasoning agents}.
The question of whether a compact model can learn high-quality representations for both text and time series, without catastrophic interference, has not been addressed.
Evidence from \citet{tan2024llmts}, who showed that LLMs do not meaningfully improve forecasting as backbones, and \citet{merrill2024language}, who found that LLMs struggle with text-encoded series, suggests that a different training paradigm may be required.
 
We present \textbf{\modelname{}}, a 324M-parameter decoder-only transformer trained from scratch on natural language and time series within a single architecture.
Both modalities share the same transformer blocks, attention mechanism, and residual stream; modality-specific components are limited to the input and output interfaces.
The bulk of pretraining uses \emph{unimodal} batches: each micro-batch contains either text tokens or time series patches, and the two modalities shape the backbone only through the shared parameters they both update.
A short second stage at extended context length introduces a small fraction of interleaved text and time series sequences for explicit cross-modal alignment.
At inference, text embeddings and patch embeddings can be freely interleaved within a single sequence, and cross-modal information flow arises naturally from causal self-attention without any architectural changes.
Unlike prior work, our goal is not conversational reasoning but learning \emph{general-purpose representations} that serve forecasting, classification, and embedding extraction, while retaining language understanding as a first-class capability.
We evaluate \modelname{} against dedicated foundation models in \emph{each} modality on \emph{their own benchmarks}: 19~NLU tasks against GPT-2 through LLaMA-3.2-1B; GIFT-Eval against the full public leaderboard; 14~UCR/UEA datasets against supervised models and frozen TSFM embeddings; and multimodal classification and forecasting on TimeCAP~\citep{gao2024timecap} and Time-MMD~\citep{liu2024timemmd}, following the MM-TSFlib~\citep{liu2024timemmd} fusion evaluation protocol.
Our contributions are:
\begin{enumerate}[leftmargin=*,itemsep=1pt,topsep=2pt,parsep=0pt]
\item \textbf{Joint pretraining from scratch.} To our knowledge, \modelname{} is the first model to learn text and time series end-to-end from random initialization within a single shared transformer backbone, rather than adapting a pretrained LLM post hoc.
\item \textbf{Cross-domain evaluation against unimodal foundation models.} We benchmark \modelname{} against scale-matched LLMs on 19 NLU tasks \emph{and} dedicated TSFMs on GIFT-Eval and UCR/UEA, addressing a longstanding gap in the multimodal time series literature.
\item \textbf{Strong frozen-backbone downstream performance.} Without per-dataset retraining, \modelname{} sets a new bar for frozen-embedding TS classification, beats every supervised fusion baseline on Time-MMD, and matches Gemma-3-270M-PT on language understanding.
\end{enumerate}

\section{Related Work}
\label{sec:related}

We situate \modelname{} within three lines of work and summarize the
architectural landscape in Table~\ref{tab:positioning}; a comprehensive
discussion appears in Appendix~\ref{app:related_full}.

\begin{table*}[b]
\vspace{-4mm}
\centering
\caption{\textbf{Positioning of \modelname{} relative to prior multimodal text and time series models.}
\emph{Base model}: the pretrained backbone (``---'' = trained from scratch).
\modelname{} is the only model that trains from scratch on both modalities and evaluates against unimodal foundation models in both domains.}
\label{tab:positioning}
\small
\setlength{\tabcolsep}{3pt}
\resizebox{\textwidth}{!}{%
\begin{tabular}{l r l l l l c c c}
\toprule
\textbf{Model}
  & \textbf{Params}
  & \textbf{Base model}
  & \textbf{TS input}
  & \textbf{Adaptation}
  & \textbf{Primary goal}
  & \textbf{\makecell{Eval vs.\\TSFMs}}
  & \textbf{\makecell{Eval vs.\\LLMs}}
  & \textbf{\makecell{MM\\eval}} \\
\midrule
LLMTIME~\citep{gruver2023llmtime}
  & 7--175B & GPT-3 / LLaMA & Digits & Frozen & Forecasting & Partial & \ding{55} & \ding{55} \\
GPT4MTS~\citep{jia2024gpt4mts}
  & $>$7B & GPT-3.5 / GPT-4 & Digits + prompt & Frozen & Forecasting & \ding{55} & \ding{55} & \ding{51} \\
Time-LLM~\citep{jin2024timellm}
  & 7B & LLaMA-7B & Patch $\to$ text proto. & Adapter & Downstream Adaptation & Partial & \ding{55} & \ding{55} \\
GPT4TS~\citep{zhou2023gpt4ts}
  & 124M & GPT-2 & Patch & Adapter (norm) & Downstream Adaptation & \ding{55} & \ding{55} & \ding{55} \\
MoAT~\citep{lee2024moat}
  & Varies & Separate enc. & Patch (decomposed) & Late fusion & Downstream Adaptation & \ding{55} & \ding{55} & \ding{51} \\
TaTs~\citep{li2025tats}
  & Varies & Separate enc. & Patch (+ text var.) & Late fusion & Downstream Adaptation & \ding{55} & \ding{55} & \ding{51} \\
\midrule
ChatTS~\citep{xie2024chatts}
  & 14B & Qwen2.5-14B & Patch (MLP enc.) & Full FT & Reasoning & \ding{55} & \ding{55} & \ding{51} \\
ChatTime~\citep{wang2024chattime}
  & 7B & LLaMA-2 & Scalar (discretized) & Full FT & Reasoning & Partial & \ding{55} & \ding{51} \\
Chat-TS~\citep{quinlan2026chattsenhancingmultimodalreasoning}
  & 8B & LLaMA-3-8B & Vocab expansion & Full FT & Reasoning & \ding{55} & \ding{51} & \ding{51} \\
MSE-ITT~\citep{koval2025msitt}
  & 8B & LLaMA-3-8B & Patch (MoE experts) & Full FT + MoE & Reasoning & \ding{55} & \ding{55} & \ding{51} \\
\midrule
\textbf{\modelname{} (ours)}
  & \textbf{324M} & \textbf{---} & \textbf{Patch (interleaved)} & \textbf{Joint (from scratch)} & \textbf{Foundation model} & \ding{51} & \ding{51} & \ding{51} \\
\bottomrule
\end{tabular}}
\end{table*}

\textbf{Time series foundation models.}
TSFMs target zero-shot generalization across domains~\citep{liang2024foundation}.
Recent models span scalar tokenization (Chronos~\citep{ansari2024chronos},
Chronos-2~\citep{ansari2025chronos2}) and patch-based encoding
(PatchTST~\citep{nie2023patchtst}, TimesFM~\citep{das2024timesfm},
PatchTST-FM~\citep{wen2026patchtst-fm}, Moirai~\citep{woo2024moirai},
MOMENT~\citep{goswami2024moment}, UniTS~\citep{gao2024units}). We compare
against these models on GIFT-Eval and UCR.

\textbf{Multimodal text and time series models.}
Prior work falls into two categories (Table~\ref{tab:positioning}).
\emph{Reasoning-focused} models including ChatTS~\citep{xie2024chatts},
Chat-TS~\citep{quinlan2026chattsenhancingmultimodalreasoning}, and
MSE-ITT~\citep{koval2025msitt} fine-tune large backbones (8--14B) on
synthetic QA data and target conversational benchmarks rather than
standard forecasting or classification.
ChatTime~\citep{wang2024chattime} instruction-tunes LLaMA-2 with
discretized series but does not evaluate on the full GIFT-Eval suite or
against scale-matched LLMs. \emph{Forecasting-focused} models fuse text
with time series via frozen LLMs (LLMTIME~\citep{gruver2023llmtime},
GPT4MTS~\citep{jia2024gpt4mts}), adapters
(Time-LLM~\citep{jin2024timellm}, GPT4TS~\citep{zhou2023gpt4ts}), or
late-fusion heads (MoAT~\citep{lee2024moat}, TaTs~\citep{li2025tats}).
Time-MMD~\citep{liu2024timemmd} contributes both a benchmark and the
MM-TSFlib fusion library, which we adopt as our multimodal baseline
protocol.

Across both categories, no prior work evaluates against dedicated TSFMs
\emph{and} dedicated LLMs on their respective benchmarks.
\citet{tan2024llmts} showed that LLM pretraining does not transfer to
forecasting, and \citet{merrill2024language} found that LLMs struggle
with text-encoded series, motivating our modality-native joint training.
\modelname{} differs from all prior work in three respects: training
from scratch on both modalities, using a compact 324M backbone focused
on representation quality rather than dialogue, and evaluating against
unimodal foundation models in both domains.

\textbf{Small language models.}
GPT-2~\citep{radford2019gpt2} demonstrated that decoder-only
transformers produce capable few-shot learners; subsequent compact
models (Qwen2~\citep{yang2024qwen2}, LLaMA-3.2~\citep{grattafiori2024llama3},
Gemma-3~\citep{team2025gemma3}, LFM-2~\citep{liquid2025lfm2}) push
zero-shot understanding to strong levels at sub-1B scale. We compare
against five such models on 19 NLU tasks to verify that joint training
preserves language capability.

\section{Methodology}
\label{sec:method}

\begin{figure*}[t]
  \centering
  \includegraphics[width=0.9\textwidth]{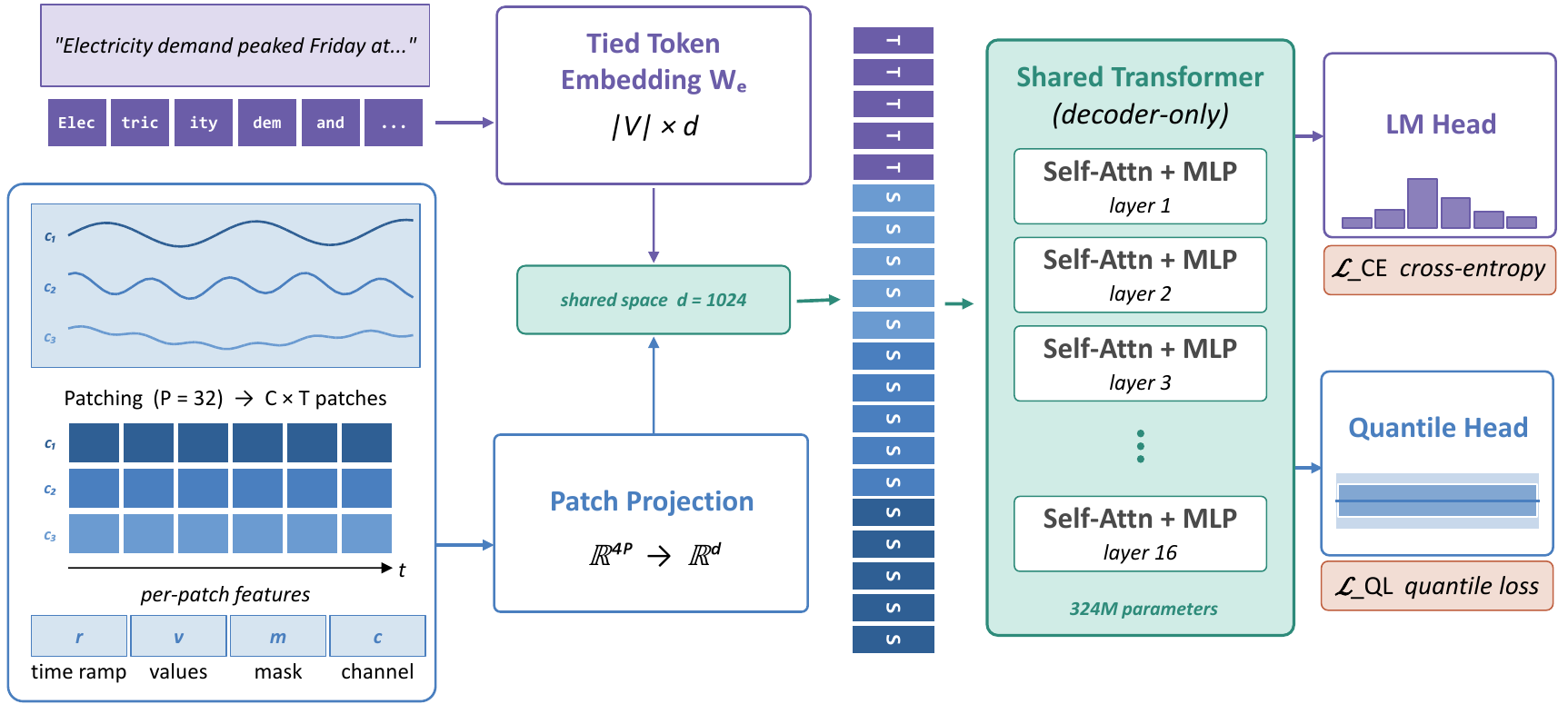}
  \caption{ \textbf{The \modelname{} architecture.} Text tokens and time series
  patches share a 16-layer decoder-only transformer, modality-specific components are limited to the
  input and output interfaces.  
  Modality-specific output heads produce quantile forecasts ($\mathcal{L}_{\mathrm{QL}}$)
  and next-token predictions ($\mathcal{L}_{\mathrm{CE}}$); the same
  backbone produces frozen embeddings for downstream classification.
  }
  \label{fig:architecture}
  \vspace{-2mm}
\end{figure*}

\modelname{} departs from prior multimodal text-and-time-series work: rather than adapting a pretrained LLM post hoc, we design an architecture in which both modalities shape a shared backbone from random initialization. The result is deliberately minimal—a decoder-only transformer~\citep{vaswani2017attention, radford2019gpt2} in which text tokens and time series patches occupy positions in a single sequence and flow through the same blocks, attention mechanism, and residual stream. Modality-specific components are confined to the interfaces: a text embedding table over a 131{,}072-entry BPE vocabulary and a patch projection on the input side; a tied language model head and a quantile head over $Q{=}21$ levels on the output side. Almost all parameters are therefore exercised by both modalities, and cross-modal information flow arises naturally from causal self-attention with no architectural additions.

\subsection{Time Series Representation}
\label{sec:ts_repr}

Following PatchTST-FM~\citep{wen2026patchtst-fm} and
Chronos-2~\citep{ansari2025chronos2}, we first standardize each input
using statistics computed \emph{only} over visible (non-NaN, unmasked)
values, then apply the inverse hyperbolic sine transform to suppress
outliers while preserving sign:
\begin{equation}
  x_{\mathrm{norm}} = \operatorname{arcsinh}\!\Bigl(\frac{x - \mu_{\mathrm{vis}}}{\sigma_{\mathrm{vis}}}\Bigr),
  \label{eq:norm}
\end{equation}
where $\mu_{\mathrm{vis}}$ and $\sigma_{\mathrm{vis}}$ are the
visible-value mean and standard deviation. During autoregressive
inference these statistics are computed once from the original context
and cached, preventing distribution drift as model-generated predictions
accumulate. The normalized series is then segmented into
$T = \lceil L / P \rceil$ non-overlapping patches of length $P{=}32$;
patching reduces effective sequence length by a factor of $P$ and gives
each token access to local temporal structure, matching the de facto
input format of recent TSFMs~\citep{nie2023patchtst, ansari2025chronos2,
liu2026moirai20timeseries}.

Each patch is represented by a $4P$-dimensional feature vector
$\mathbf{f} = [\mathbf{r}; \mathbf{v}; \mathbf{m}; \mathbf{c}]$ obtained
by concatenating four $P$-dimensional components: a \emph{time ramp}
$\mathbf{r}$ encoding the patch's normalized position within its channel
(running from approximately $-1$ at the start of a channel to $0$ at its
end and resetting at channel boundaries for multivariate inputs); the
\emph{normalized values} $\mathbf{v}$ produced by Eq.~\ref{eq:norm}; a
binary \emph{validity mask} $\mathbf{m} \in \{0,1\}^P$ that distinguishes
observed values from missing or masked positions; and a \emph{channel ramp} $\mathbf{c}$ that encodes channel identity in
multivariate inputs. For a sample with $C$ channels, channel $j \in
\{0,\ldots,C-1\}$ is assigned the scalar value
$j/\max(C-1,1)$, which is expanded across all $P$ positions of each
patch from that channel. Thus multivariate channel values are evenly
spaced in $[0,1]$, while univariate inputs use $\mathbf{c}=\mathbf{0}$.
If channel identifiers are unavailable, the channel ramp defaults to
zero.
Appendix~\ref{app:channel_ramp} shows that channel-aware multivariate
handling improves classification performance over
mean-channel pooling on average on the multivariate UEA datasets.
The feature vector is projected to the transformer embedding dimension
$d$ via a single bias-free linear layer followed by RMSNorm, and the resulting patch embedding is placed into the shared input space.
\begin{equation}
  \mathbf{e}_{\mathrm{ts}} = \mathrm{RMSNorm}(\mathbf{W}_p \mathbf{f}) \in \mathbb{R}^d,
  \label{eq:patchproj}
\end{equation}

Text tokens are embedded via a learned table $\mathbf{W}_e \in
\mathbb{R}^{|\mathcal{V}| \times d}$ that is tied with the language
model output head~\citep{press-wolf-2017-using}, an important saving
given the 131k-entry vocabulary. During pretraining, batches are
either text-only or time-series-only (with the small interleaved
fraction in stage 2 described in Section~\ref{sec:implementation}).

\subsection{Output Heads and Training Objective}

At text positions, transformer hidden states are projected to vocabulary
logits via the tied embedding matrix, with logit
soft-capping~\citep{team2025gemma3}
$\ell \leftarrow \alpha \tanh(\ell/\alpha)$, $\alpha{=}15$, applied to
prevent extreme pre-softmax values. The text loss is standard
autoregressive cross-entropy $\mathcal{L}_{\mathrm{CE}}$. At time series
positions, RMSNorm-projected hidden states are mapped through a single
bias-free linear layer to $P \times Q$ outputs, where $Q{=}21$ quantile
levels are spaced uniformly over $\tau \in [0.05, 0.95]$. We minimize
the masked quantile loss~\citep{gneiting2007crps}:
\begin{equation}
  \mathcal{L}_{\mathrm{QL}} =
  \frac{
  \sum_{b,t,p,q} z_{b,t,p}\,
  \rho_{\tau_q}\!\bigl(y_{b,t,p} - \hat{q}_{b,t,p,q}\bigr)
  }{
  Q \sum_{b,t,p} z_{b,t,p}
  },
  \quad
  \rho_\tau(u) = \max\!\bigl(\tau u,\,(\tau-1)u\bigr),
  \label{eq:qloss}
\end{equation}
where $z_{b,t,p} \in \{0,1\}$ is a target-validity mask that is one only
for finite, observed target values inside the prediction horizon and
zero for padded or otherwise invalid positions. The mask is applied to
each per-position quantile loss term before normalization, so padded
targets in partial forecast patches do not contribute to the objective.

At inference time, we denormalize predictions by inverting
Eq.~\ref{eq:norm}:
$\hat{x} = \sinh(\hat{q}) \cdot \sigma_{\mathrm{vis}} + \mu_{\mathrm{vis}}$.

The overall training objective is the weighted sum
$\mathcal{L} = w_{\mathrm{text}} \mathcal{L}_{\mathrm{CE}} +
w_{\mathrm{TS}} \mathcal{L}_{\mathrm{QL}}$ with $w_{\mathrm{text}}{=}1.0$
and $w_{\mathrm{TS}}{=}2.5$; the asymmetric weighting reflects the
substantially smaller scale of the per-element quantile loss relative
to cross-entropy and balances gradient contributions from the two
modalities.

\section{Implementation}
\label{sec:implementation}
\modelname{} is a 16-layer, 324M-parameter decoder-only transformer ($d{=}1024$, 8 GQA heads with 4 KV heads, RoPE~\citep{su2024rope}, SwiGLU~\citep{shazeer2020gluvariantsimprovetransformer}, pre-norm RMSNorm~\citep{zhang2019rmsnorm}), with patch length $P{=}32$ and a $Q{=}21$-quantile head over $\tau \in [0.05, 0.95]$. Pretraining runs on $2{\times}$H100 80GB GPUs in BF16. \textbf{Stage 1} trains at sequence length $2048$ for $47{,}683$ steps (${\sim}3.1$M tokens/batch), yielding ${\sim}138$B text and ${\sim}12$B TS patches; each micro-batch is text-only ($p{=}0.92$, from FineWeb-Edu~\citep{penedo2024fineweb} and Dolmino-mix-1124~\citep{olmo2024dolmino}) or time-series-only ($p{=}0.08$, from GiftEvalPretrain plus KernelSynth augmentation; Appendix~\ref{app:synthetic}). \textbf{Stage 2} extends context to $4096$ and replaces $5\%$ of TS tokens with interleaved alignment data from ChatTS~\citep{xie2024chatts} and \citet{merrill2024language}, establishing cross-modal correspondences while preserving stage-1 capabilities. The text-heavy 92/8 mix is a compute constraint: matching text-only baselines trained on trillions of tokens requires devoting most of our budget to text (Section~\ref{sec:eval_pretrain_nlu}). Full details appear in Appendix~\ref{app:impl}.

\section{Evaluation}
\label{sec:experiments}

\begin{wraptable}{r}{0.58\textwidth}
\vspace{-12mm}
\vspace{-\intextsep}
\centering
\caption{Language understanding (19 NLU tasks). Models ordered by parameter count. Subscripts indicate shot count. Chr.=Chronicle model.}
\label{tab:nlu_full}
\scriptsize
\setlength{\tabcolsep}{1pt}
\begin{tabular}{l ccccccc}
\toprule
\textbf{Task}
  & \textbf{GPT-2}
  & \textbf{Gemma-3}
  & \textbf{Chr.-2}
  & \textbf{Chr.-1}
  & \textbf{LFM-2}
  & \textbf{Qwen2}
  & \textbf{LLaMA-3.2} \\
\textit{Params}
  & \textit{124M} & \textit{270M} & \textit{324M}
  & \textit{324M}
  & \textit{350M} & \textit{500M} & \textit{1.2B} \\
\textit{Tokens}
  & \textit{${\sim}10$B} & \textit{6T} & \textit{${\sim}153$B}
  & \textit{${\sim}138$B}
  & \textit{10T} & \textit{12T} & \textit{9T} \\
\midrule
HellaSwag$_{0}$    & 0.310 & 0.401 & 0.435 & 0.430 & 0.483 & 0.480 & \textbf{0.629} \\
HellaSwag$_{10}$   & 0.308 & 0.397 & 0.429 & 0.427 & 0.473 & 0.482 & \textbf{0.648} \\
ARC-E$_{10}$       & 0.417 & 0.583 & 0.651 & 0.644 & \textbf{0.715} & 0.595 & 0.678 \\
ARC-C$_{10}$       & 0.224 & 0.289 & 0.325 & 0.325 & \textbf{0.445} & 0.311 & 0.376 \\
COPA$_{0}$         & 0.630 & 0.670 & 0.660 & 0.670 & 0.690 & 0.670 & \textbf{0.760} \\
CSQA$_{10}$        & 0.230 & 0.207 & 0.193 & 0.235 & 0.541 & \textbf{0.582} & 0.370 \\
PiQA$_{10}$        & 0.624 & 0.676 & 0.694 & 0.684 & 0.698 & 0.700 & \textbf{0.757} \\
LAMBADA$_{0}$      & 0.322 & 0.429 & 0.382 & 0.397 & 0.398 & 0.494 & \textbf{0.627} \\
Winograd$_{0}$     & 0.575 & 0.652 & 0.663 & 0.641 & 0.608 & 0.696 & \textbf{0.799} \\
WinoGrande$_{0}$   & 0.507 & 0.536 & 0.516 & 0.509 & 0.558 & 0.557 & \textbf{0.609} \\
BoolQ$_{10}$       & 0.547 & 0.517 & 0.566 & 0.570 & 0.572 & 0.614 & \textbf{0.657} \\
CoQA$_{0}$         & 0.136 & 0.223 & 0.211 & 0.221 & 0.303 & 0.324 & \textbf{0.360} \\
SQuAD$_{10}$       & 0.058 & 0.250 & 0.290 & 0.300 & 0.318 & \textbf{0.492} & 0.479 \\
Jeopardy$_{10}$    & 0.003 & 0.130 & 0.101 & 0.121 & 0.069 & 0.139 & \textbf{0.344} \\
BB WikiQA$_{10}$   & 0.283 & 0.548 & 0.538 & 0.533 & 0.406 & 0.594 & \textbf{0.643} \\
BB CSAlg$_{10}$    & 0.423 & 0.436 & 0.405 & 0.389 & 0.405 & 0.442 & \textbf{0.458} \\
BB Ops$_{10}$      & 0.090 & 0.210 & 0.167 & 0.176 & 0.319 & 0.305 & \textbf{0.405} \\
AGIE LSAT$_{3}$    & 0.209 & \textbf{0.300} & 0.239 & 0.283 & 0.243 & 0.252 & 0.239 \\
BB LangID$_{10}$   & 0.258 & 0.254 & 0.248 & 0.248 & 0.282 & \textbf{0.318} & 0.253 \\
\midrule
\textbf{Average}   & 0.324 & 0.406 & 0.406 & 0.411 & 0.449 & 0.476 & \textbf{0.531} \\
\bottomrule
\end{tabular}
\vspace{-12mm}
\end{wraptable}

We evaluate \modelname{} across five benchmarks organized into two
tiers. \emph{Pretraining benchmarks}
(Sections~\ref{sec:eval_pretrain_nlu} and~\ref{sec:eval_pretrain_gift})
probe each training objective in isolation, measuring whether text
capability survives joint training and how well zero-shot forecasting
generalizes across domains. \emph{Downstream application benchmarks}
(Section~\ref{sec:eval_downstream}) assess
whether the learned representations transfer to downstream multi-modal and uni-modal tasks.

\subsection{Pretraining Effectiveness}
\label{sec:eval_pretrain}
 
\subsubsection{Language Understanding}
\label{sec:eval_pretrain_nlu}

We evaluate on 19 NLU tasks drawn from the DCLM evaluation
suite~\citep{li2024dclm}, spanning commonsense reasoning
(HellaSwag~\citep{zellers2019hellaswag},
COPA~\citep{roemmele2011copa},
PiQA~\citep{bisk2020piqa},
CommonsenseQA~\citep{talmor2019commonsenseqa},
WinoGrande~\citep{sakaguchi2021winogrande}),
reading comprehension
(ARC-Easy, ARC-Challenge~\citep{clark2018arc},
BoolQ~\citep{clark2019boolq},
CoQA~\citep{reddy2019coqa},
SQuAD~\citep{rajpurkar2016squad}),
cloze and completion
(LAMBADA~\citep{paperno2016lambada},
Winograd~\citep{levesque2012winograd}),
knowledge
(Jeopardy, BB~WikiQA~\citep{srivastava2023bigbench}),
and algorithmic reasoning
(BB~CS~Algorithms, BB~Operators, BB~Language~ID~\citep{srivastava2023bigbench};
AGI-Eval LSAT~\citep{zhong2023agieval}).
All tasks use zero-shot or few-shot in-context learning with no
fine-tuning. We compare against five text-only decoder-only language
models: GPT-2~\citep{radford2019gpt2} (124M),
Gemma-3-270M-PT~\citep{team2025gemma3} (270M),
LFM-2-350M~\citep{liquid2025lfm2} (350M),
Qwen2-0.5B~\citep{yang2024qwen2} (500M), and
LLaMA-3.2-1B~\citep{grattafiori2024llama3} (1.2B).
 
Table~\ref{tab:nlu_full} reports all 19 tasks.
Stage~1 achieves an average accuracy of $0.411$ and Stage~2 achieves $0.406$,
with Stage~2 matching Gemma-3-270M-PT ($0.406$) at comparable scale; both
stages sit between GPT-2 ($0.324$) and Qwen2-0.5B ($0.476$).
The small advantage of Stage~1 is consistent with Stage~2 replacing a fraction
of text tokens with multimodal alignment data, slightly reducing the effective
language training budget, and mirrors the pattern observed across all unimodal
benchmarks.
On ARC-Easy, Stage~2 ($0.651$) closely approaches LLaMA-3.2-1B ($0.678$),
a model roughly $4\times$ larger trained exclusively on text, and outperforms
both Gemma-3-270M-PT ($0.583$) and Qwen2-0.5B ($0.595$).
Both stages match or exceed GPT-2 on the vast majority of tasks.
The training budget context is an important consideration: \modelname{} sees only ${\sim}138$B text tokens during pretraining, roughly $43\times$ fewer than Gemma-3-270M (6T), $72\times$ fewer than LFM-2-350M (10T), and $87\times$
fewer than Qwen2-0.5B (12T).
These results demonstrate that devoting ${\sim}8\%$ of training compute to
time series does not cause catastrophic interference in the shared transformer
backbone, and validate the text-heavy token mix described in
Section~\ref{sec:implementation}.

\subsubsection{Zero-Shot Probabilistic Forecasting}
\label{sec:eval_pretrain_gift}

GIFT-Eval~\citep{jain2024gifteval} comprises 97 zero-shot forecasting
tasks drawn from 55 datasets across 7 domains at three horizon lengths
(short, medium, long). All metrics are standardized by dividing by the
Seasonal Naive baseline and aggregated via geometric mean; we report
MASE (point forecast quality using the median quantile) and WQL
(weighted quantile loss, equivalent to CRPS). We compare against
published scores from the public GIFT-Eval leaderboard, representing
the strongest dedicated TSFMs currently evaluated: the leading
zero-shot models
PatchTST-FM-r1~\citep{wen2026patchtst-fm},
TimesFM-2.5~\citep{das2024timesfm},
TiRex~\citep{auer2025tirex},
Toto-Base~\citep{cohen2025toto},
YingLong-300M~\citep{wang2025yinglong}, Chronos-2-Synth~\citep{ansari2025chronos2}, and
Moirai-Large~\citep{woo2024moirai};
supervised baselines PatchTST~\citep{nie2023patchtst},
N-BEATS~\citep{oreshkin2019nbeats},
DLinear~\citep{zeng2023dlinear}, and
DeepAR~\citep{salinas2020deepar};
and statistical baselines Seasonal Naive and Auto-ARIMA. We exclude models trained with potentially leaky data from test-set distributions.
\modelname{} forecasts autoregressively, generating one patch per step.

\begin{figure}[!t]
  \centering
  \includegraphics[width=0.9\textwidth]{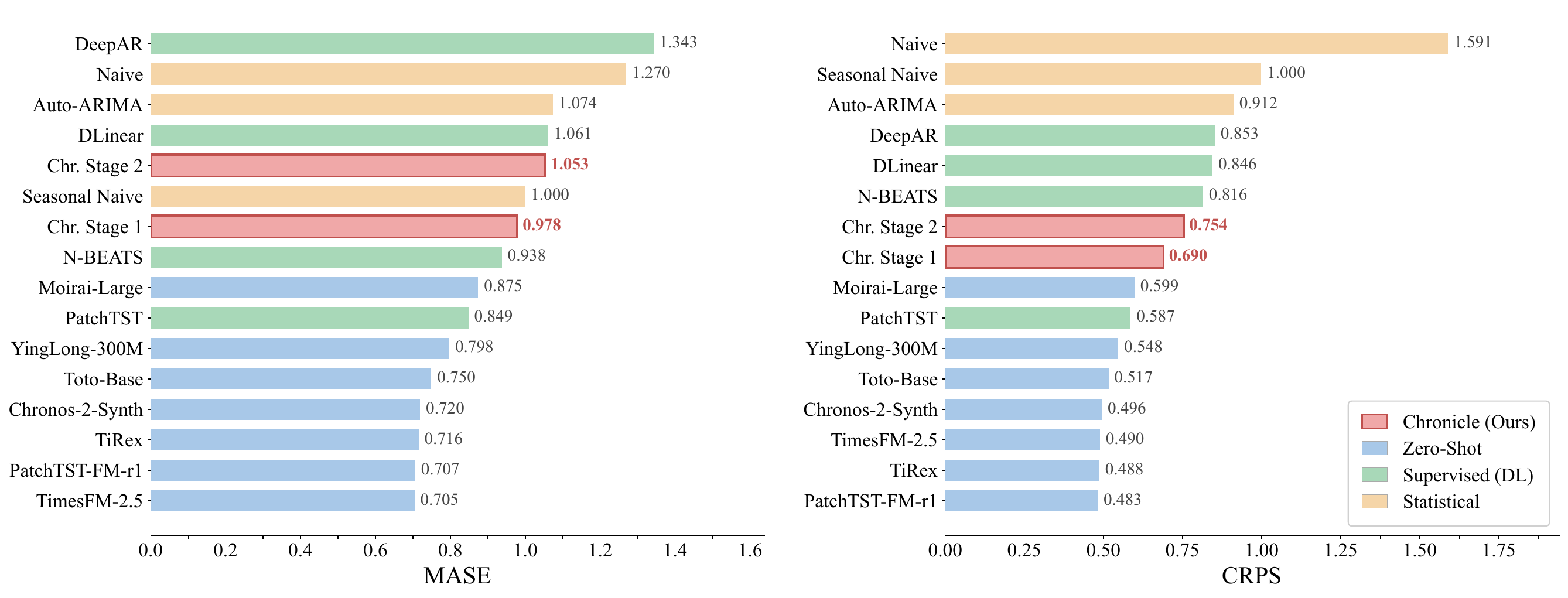}
  \caption{\textbf{GIFT-Eval leaderboard (97 tasks; lower is better).}
    MASE (left) and CRPS (right) for comparative models, plus \modelname{} Stage~1 and Stage~2
    (highlighted). Stage~1 is the stronger pure forecaster, while Stage~2
    is the aligned checkpoint used for multimodal transfer.}
  \label{fig:gifteval_leaderboard}
  \vspace{-4mm}
\end{figure}

Figure~\ref{fig:gifteval_leaderboard} places both \modelname{} checkpoints
within the full leaderboard. Stage~1 is the stronger pure zero-shot forecaster,
reaching $0.978$ MASE and $0.690$ CRPS, while Stage~2 reaches $1.053$ MASE
and $0.754$ CRPS after the alignment stage. This establishes the main tradeoff:
unimodal training gives the best isolated forecasting performance, whereas
adding a small fraction of interleaved text and time-series data slightly reduces
GIFT-Eval scores but improves downstream multimodal transfer. Despite allocating
only ${\sim}8\%$ of training compute to time series, Stage~1 outperforms
Seasonal Naive on both metrics and improves over several supervised and
statistical baselines on CRPS, including N-BEATS ($0.816$), DLinear ($0.846$),
DeepAR ($0.853$) and Auto-ARIMA ($0.912$).
The remaining gap to dedicated TSFMs reflects two principled design choices:
(i) only ${\sim}8\%$ of training compute is allocated to time series versus
$100\%$ for dedicated models; and (ii) to align with our text setup we use
causal next-patch prediction, while PatchTST-FM uses contiguous patch masking
in an otherwise similar architecture, which their ablations show meaningfully
reduces MASE.

\subsection{Downstream Applications}
\label{sec:eval_downstream}

We now evaluate whether \modelname{}'s learned representations transfer
to three downstream tasks: multimodal classification, multimodal
forecasting, and time series classification. All three tasks probe
different aspects of the model's representations (cross-modal fusion,
text-conditioned prediction, and temporal discriminability) without
retraining the backbone.

\paragraph{Shared baselines.}
All downstream evaluations draw from a common pool of baselines.
\emph{Supervised DL models} (Informer~\citep{zhou2021informer},
TimesNet~\citep{wu2023timesnet}, Autoformer~\citep{wu2021autoformer},
iTransformer~\citep{liu2024itransformer}, DLinear~\citep{zeng2023dlinear},
PatchTST~\citep{nie2023patchtst}, and FEDformer~\citep{zhou2022fedformer})
are trained independently per dataset for TS classification.
\emph{TS foundation models} (Chronos-2~\citep{ansari2025chronos2},
Moirai-2~\citep{liu2026moirai20timeseries}, and
TimesFM~\citep{das2024timesfm}) are evaluated with a learned linear
probe on frozen embeddings for classification and via fusion heads or
direct prediction for forecasting.

\emph{Multimodal fusion baselines} follow the
\textbf{MM-TSFlib}~\citep{liu2024timemmd} protocol, the standard fusion
library introduced alongside the Time-MMD benchmark and subsequently
adopted by multiple text-augmented time series studies. Under this
protocol, each baseline pairs a \emph{trainable} time series encoder
(DLinear, PatchTST, or TimesNet) with a \emph{frozen} pretrained text
encoder (BERT~\citep{devlin2019bert} or GPT-2~\citep{radford2019gpt2})
and a trainable two-layer MLP fusion head. The TS encoder is fine-tuned
end-to-end on each dataset together with the head, so the temporal
representation adapts to the task. We additionally report \emph{FM
Fusion} baselines that pair the same frozen text encoders with frozen
TS foundation models (Chronos-2, Moirai-2, TimesFM) as encoders and
train only the fusion head.

\subsubsection{Multimodal Classification}
\label{sec:eval_mm_class}
\begin{wraptable}{rb}{0.55\textwidth}
\centering
\vspace{-12mm}
\vspace{-\intextsep}
\caption{Multimodal classification on TimeCAP. Scores are averaged over Weather, Finance, and Healthcare. Values are mean $\pm$ standard deviation over 3 seeds. Chronicle rows report both LP and LoRA scores with TS token repeat $r\in\{1,64\}$.}
\label{tab:mm-cls-main}
\scriptsize
\setlength{\tabcolsep}{2pt}
\begin{tabular}{l l c c}
\toprule
\textbf{Category} & \textbf{Model} & {\textbf{F1}\,$\uparrow$} & {\textbf{AUC}\,$\uparrow$} \\
\midrule
\multirow{6}{*}{MM-TSFlib} & DLin+BERT & $0.588 \pm 0.016$ & $0.739 \pm 0.024$ \\
 & DLin+GPT2 & $0.564 \pm 0.026$ & $0.724 \pm 0.017$ \\
 & PTST+BERT & $0.578 \pm 0.022$ & $0.719 \pm 0.022$ \\
 & PTST+GPT2 & $0.539 \pm 0.021$ & $0.707 \pm 0.034$ \\
 & TNet+BERT & $0.589 \pm 0.021$ & $0.750 \pm 0.026$ \\
 & TNet+GPT2 & $0.577 \pm 0.019$ & $0.754 \pm 0.028$ \\
\midrule
\multirow{6}{*}{FM Fusion} & BERT+Chr2 & $0.590 \pm 0.021$ & $0.726 \pm 0.023$ \\
 & BERT+Moi2 & $0.588 \pm 0.004$ & $0.751 \pm 0.025$ \\
 & BERT+TFM & $0.498 \pm 0.006$ & $0.659 \pm 0.023$ \\
 & GPT2+Chr2 & $0.455 \pm 0.043$ & $0.673 \pm 0.056$ \\
 & GPT2+Moi2 & $0.542 \pm 0.038$ & $0.739 \pm 0.015$ \\
 & GPT2+TFM & $0.480 \pm 0.034$ & $0.628 \pm 0.018$ \\
\midrule
\multirow{8}{*}{\textbf{\modelname{}}} & Stage 1 LP (r=1) & $0.593 \pm 0.021$ & $0.733 \pm 0.030$ \\
 & Stage 1 LP (r=64) & $0.608 \pm 0.010$ & $0.745 \pm 0.014$ \\
 & Stage 1 LoRA (r=1) & $0.601 \pm 0.011$ & $0.739 \pm 0.017$ \\
 & Stage 1 LoRA (r=64) & $0.584 \pm 0.024$ & \textbf{\boldmath $0.763 \pm 0.011$} \\
 & Stage 2 LP (r=1) & $0.594 \pm 0.025$ & $0.731 \pm 0.029$ \\
 & Stage 2 LP (r=64) & $0.605 \pm 0.011$ & $0.750 \pm 0.014$ \\
 & Stage 2 LoRA (r=1) & $0.595 \pm 0.030$ & $0.731 \pm 0.032$ \\
 & Stage 2 LoRA (r=64) & \textbf{\boldmath $0.613 \pm 0.011$} & $0.757 \pm 0.014$ \\
\bottomrule
\end{tabular}
\vspace{-2mm}
\end{wraptable}

We evaluate multimodal classification on TimeCAP~\citep{gao2024timecap}
across three domains: \emph{Weather} (binary rain/no-rain),
\emph{Finance} (three-way market direction), and \emph{Healthcare}
(mean of two binary tasks: in-hospital mortality and disease test-positive prediction).
All four underlying tasks are class-imbalanced (majority class 61--69\%);
we train every method with class-balanced cross-entropy, cap text inputs at $384$ tokens,
and report macro-F1 and AUC. \modelname{} adds a 2-layer head on a single fully frozen
backbone with joint text+TS input; full training settings appear in
Appendix~\ref{app:eval_timecap} and we follow the training splits from~\citet{gao2024timecap}.
Table~\ref{tab:mm-cls-main} reports results averaged over Weather, Finance, and Healthcare.
Several baselines achieve inflated accuracy via majority-class collapse but perform
poorly on these imbalance-aware metrics. Across macro-F1 and AUC, \modelname{} is the
strongest entry: Stage~2 LoRA ($r{=}64$) achieves the best average macro-F1
($\mathbf{0.613}$), and Stage~1 LoRA ($r{=}64$) achieves the best average AUC
($\mathbf{0.763}$), both with tight variance across seeds. The best MM-TSFlib
and FM Fusion baselines reach macro-F1 of $0.590$ (TNet+BERT) and AUC of
$0.754$ (TNet+GPT2), respectively, trailing \modelname{} on both metrics.

\subsubsection{Multimodal Forecasting}
\label{sec:eval_mm_forecast}

We evaluate multimodal forecasting on the 9 Time-MMD~\citep{liu2024timemmd}
domains (agriculture through traffic; textual fact reports;
chronological 70/10/20 splits), reporting MAE averaged over all
forecast horizons per domain (monthly: 6 to 12 steps; weekly: 12 to
48; daily: 48 to 336). For \modelname{}, we report two variants:
\emph{ZS}, zero-shot forecasting, and
\emph{LP}, where a forecasting head is trained on top of the frozen
backbone with joint text+TS input. 

Table~\ref{tab:mm_forecasting} presents per-domain MAE and normalized mean
absolute error (NMAE) to account for differing data scales.
\modelname{} Stage~2 (LP) achieves the best overall NMAE ($\mathbf{0.514}$)
and average rank ($\mathbf{2.56}$), outperforming the strongest MM-TSFlib
baseline (BERT+TNet, NMAE $0.621$, rank $8.56$) and the strongest FM Fusion
baseline (GPT2+Moi2, NMAE $0.588$, rank $6.44$).
Stage~1 (LP) also surpasses all baselines (NMAE $0.524$, rank $5.00$), and
Stage~2 improves over Stage~1 on both metrics, directly validating the
multimodal alignment stage.
At the domain level, Stage~2 (LP) leads on 5 of 9 domains
(Energy, Environment, Public Health, Security, and Social Good) and is
within $0.002$ of the best method on Agriculture and Climate.
The improvement from Stage~1 (ZS) to Stage~2 (ZS) (NMAE $1.040 \to 0.835$)
shows that even zero-shot multimodal forecasting benefits from the alignment
stage, while linear probing unlocks large additional gains
on Environment ($-0.860$ MAE), Public Health ($-0.800$), Energy ($-0.278$),
Traffic ($-0.141$), and Social Good ($-0.120$).

\begin{table*}[t]
  \centering
  \caption{\textbf{Multimodal forecasting MAE on Time-MMD (lower is better).} MM-TSFlib baselines finetune the TS encoder
  end-to-end with a frozen text encoder and a trained MLP head.
  FM Fusion baselines pair a frozen pretrained TS foundation model
  with a frozen text encoder and a trained fusion head. \modelname{} reports zero-shot and finetuned head variants. Abbreviations: Agri.=Agriculture, Clim.=Climate,
  Econ.=Economy, Enrg.=Energy, Env.=Environment, P.Hlth=Public Health,
  Sec.=Security, Soc.G=Social Good, Traf.=Traffic; DLin=DLinear,
  PTST=PatchTST, TNet=TimesNet, Chr2=Chronos-2, Moi2=Moirai-2,
  TFM=TimesFM.}
  \label{tab:mm_forecasting}
  \scriptsize
  \setlength{\tabcolsep}{3pt}
  \resizebox{0.96\textwidth}{!}{%
  \begin{tabular}{l l
    S[table-format=1.3] S[table-format=1.3] S[table-format=1.3]
    S[table-format=1.3] S[table-format=1.3] S[table-format=1.3]
    S[table-format=1.3] S[table-format=1.3] S[table-format=1.3]
    S[table-format=1.3] S[table-format=2.2]}
  \toprule
  \textbf{Cat.} & \textbf{Model}
    & {Agri.} & {Clim.} & {Econ.} & {Enrg.} & {Env.}
    & {P.Hlth} & {Sec.} & {Soc.G} & {Traf.}
    & {\textbf{NMAE}} & {\textbf{Avg Rank}} \\
  \midrule
  \multirow{6}{*}{\makecell[l]{MM-TSFlib\\(TS enc.\\trainable,\\text frozen)}}
    & BERT+DLin  & \textbf{0.181} & 0.901 & \textbf{0.069} & 0.409 & 0.447 & 0.839 & 2.067 & 0.499 & 0.241 & 0.648 &  7.44 \\
    & BERT+PTST  & 0.194 & 0.902 & 0.075 & 0.459 & 0.452 & 0.825 & 2.179 & 0.526 & 0.220 & 0.666 & 10.22 \\
    & BERT+TNet  & 0.186 & 0.913 & 0.073 & 0.437 & 0.438 & 0.818 & 1.707 & 0.535 & 0.226 & 0.621 &  8.56 \\
    & GPT2+DLin  & 0.184 & 0.890 & 0.073 & 0.424 & 0.447 & 0.790 & 1.871 & 0.523 & 0.225 & 0.626 &  6.78 \\
    & GPT2+PTST  & 0.195 & 0.884 & 0.076 & 0.449 & 0.465 & 0.794 & 1.772 & 0.562 & \textbf{0.209} & 0.627 &  9.00 \\
    & GPT2+TNet  & \textbf{0.181} & 0.924 & 0.076 & 0.464 & 0.450 & 0.833 & 1.780 & 0.591 & 0.211 & 0.640 &  9.78 \\
  \midrule
  \multirow{6}{*}{\makecell[l]{FM Fusion\\(both enc.\\frozen, head\\trained)}}
    & BERT+Chr2  & 0.211 & 0.894 & 0.103 & 0.505 & 0.493 & 1.274 & 1.700 & 0.527 & 0.386 & 0.719 & 12.78 \\
    & BERT+Moi2  & 0.185 & \textbf{0.873} & 0.073 & 0.453 & 0.507 & 0.790 & 1.413 & 0.464 & 0.227 & 0.591 &  6.89 \\
    & BERT+TFM   & 0.190 & 0.881 & 0.071 & 0.403 & 0.491 & 0.765 & 1.686 & 0.441 & 0.228 & 0.601 &  6.11 \\
    & GPT2+Chr2  & 0.213 & 0.894 & 0.103 & 0.505 & 0.494 & 1.273 & 1.616 & 0.523 & 0.386 & 0.712 & 12.22 \\
    & GPT2+Moi2  & 0.189 & 0.878 & 0.073 & 0.448 & 0.514 & 0.785 & 1.394 & 0.453 & 0.223 & 0.588 &  6.44 \\
    & GPT2+TFM   & 0.187 & 0.878 & 0.079 & 0.396 & 0.491 & 0.777 & 1.575 & 0.471 & 0.229 & 0.595 &  6.89 \\
  \midrule
  \multirow{4}{*}{\textbf{\modelname{}}}
    & Stage 1 (ZS)
      & 0.269 & 1.903 & 0.247 & 0.444 & 1.397 & 1.061 & 1.222 & 0.817 & 1.052 & 1.040 & 13.44 \\
    & Stage 1 (LP)
      & 0.184 & 0.894 & 0.074 & 0.390 & 0.417 & 0.697 & 1.058 & 0.413 & 0.247 & 0.524 &  5.00 \\
    & Stage 2 (ZS)
      & 0.222 & 0.890 & 0.100 & 0.648 & 1.270 & 1.490 & 1.082 & 0.519 & 0.386 & 0.835 & 11.89 \\
    & Stage 2 (LP)
      & \textbf{0.181} & 0.875 & 0.070 & \textbf{0.370} & \textbf{0.410}
      & \textbf{0.690} & \textbf{1.056} & \textbf{0.399} & 0.245 & \textbf{0.514} & \textbf{2.56} \\
  \bottomrule
  \vspace{-6mm}
  \end{tabular}}
  \end{table*}
  
\vspace{-2mm}
\subsection{Time-Series Classification}
\label{sec:eval_tsclass}

\begin{wraptable}{rb}{0.52\textwidth}
\centering
\vspace{-8mm}
\vspace{-\intextsep}
\caption{Time series classification on 24 UCR/UEA datasets. Linear probes on frozen embeddings for TS foundation models and Chronicle; supervised DL baselines are trained per dataset. Full results in Table~\ref{tab:ucr-uea-full}. Results are averaged over 5 different seeds.}
\label{tab:ucr-uea-main}
\scriptsize
\setlength{\tabcolsep}{2pt}
\begin{tabular}{l l c c}
\toprule
\textbf{Category} & \textbf{Model}
  & {\textbf{Acc}\,$\uparrow$}
  & {\textbf{F1}\,$\uparrow$} \\
\midrule
\multirow{7}{*}{\makecell[l]{Supervised DL}}
  & Informer        & $0.565 \pm 0.213$ & $0.483 \pm 0.241$ \\
  & TimesNet        & $0.645 \pm 0.228$ & $0.575 \pm 0.267$ \\
  & Autoformer      & $0.628 \pm 0.221$ & $0.566 \pm 0.259$ \\
  & DLinear         & $0.637 \pm 0.228$ & $0.606 \pm 0.248$ \\
  & iTransformer    & $0.628 \pm 0.221$ & $0.582 \pm 0.249$ \\
  & FEDformer       & $0.723 \pm 0.229$ & $0.666 \pm 0.283$ \\
  & PatchTST        & $0.668 \pm 0.231$ & $0.618 \pm 0.267$ \\
\midrule
\multirow{3}{*}{\makecell[l]{TS Foundation\\Models (LP)}}
  & Chronos-2       & $0.376 \pm 0.225$ & $0.230 \pm 0.167$ \\
  & TimesFM         & $0.611 \pm 0.238$ & $0.563 \pm 0.262$ \\
  & Moirai-2        & $0.714 \pm 0.238$ & $0.692 \pm 0.257$ \\
\midrule
\multirow{2}{*}{\textbf{\modelname{}}}
  & Stage 1 & \textbf{\boldmath $0.736 \pm 0.206$} & \textbf{\boldmath $0.712 \pm 0.226$} \\
  & Stage 2 & $0.729 \pm 0.199$ & $0.700 \pm 0.220$ \\
\bottomrule
\end{tabular}
\vspace{-4mm}
\end{wraptable}

We evaluate time-series classification on 24 datasets: 14 univariate
datasets from the UCR Time Series Archive~\citep{dau2019ucr} and 10
multivariate datasets from the UEA archive, using official train/test
splits. Supervised DL baselines are trained per-dataset
with Adam ($\mathrm{lr}{=}10^{-3}$, batch size 16). Foundation model
baselines and \modelname{} use a learned linear probe on frozen
embeddings, directly testing whether pretrained temporal
representations are linearly separable without backbone adaptation.

Table~\ref{tab:ucr-uea-main} shows accuracy and F1 across all 24 datasets.
\modelname{} Stage~1 achieves the strongest overall results among
frozen-backbone models, reaching 0.736 accuracy and 0.712 F1,
compared to $0.714 / 0.692$ for Moirai-2, $0.611 / 0.563$ for TimesFM,
and $0.376 / 0.230$ for Chronos-2.
Stage~2 reaches $0.729 / 0.700$, a small but consistent decrease relative
to Stage~1 that mirrors the pattern on GIFT-Eval and NLU: replacing a
portion of unimodal time series tokens with multimodal alignment data
modestly reduces purely unimodal representation quality while improving
cross-modal tasks (Section~\ref{sec:eval_mm_forecast},
Section~\ref{sec:eval_mm_class}).
Both stages exceed all supervised DL baselines trained per-dataset except
FEDformer ($0.723$ accuracy), from a single frozen backbone with no
per-dataset retraining.
The full results are given in Table~\ref{tab:ucr-uea-full}.

\subsection{TS-Token Repetition for Short Time Series}
\label{sec:tok_repeat}

\begin{figure}[t]
\centering
\includegraphics[width=0.9\textwidth]{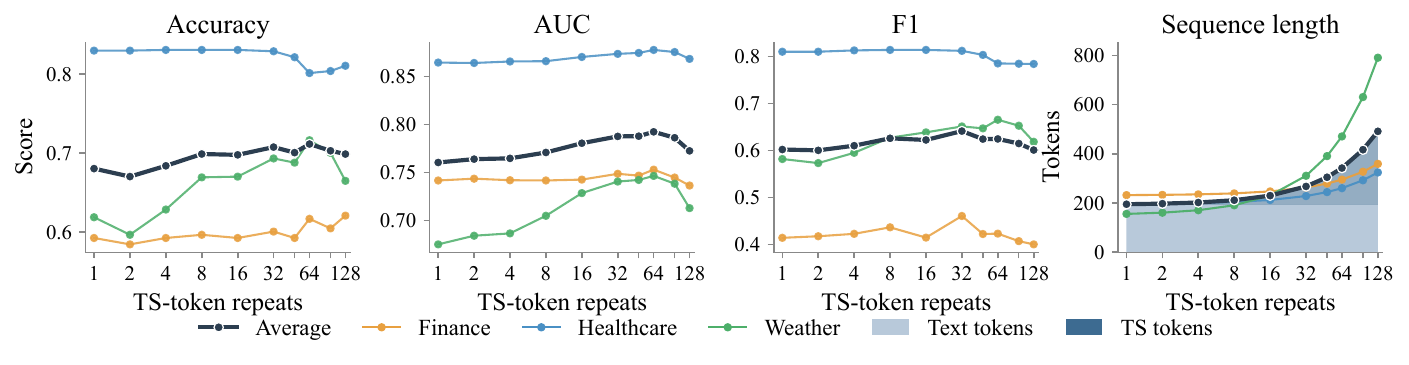}
  \caption{\textbf{Effect of TS-token repetition on multimodal
    classification.} Accuracy (left), AUC (middle), and macro-F1
    (right) as a function of TS-token repeats $r$, evaluated on the
    three TimeCAP domains and averaged (dashed black). Repetition
    rebalances the text--TS token ratio in the shared sequence;
    performance peaks near $r{=}64$ then degrades as attention
    dilutes across identical copies.}
  \label{fig:tok_repeat}
  \vspace{-4mm}
\end{figure}

Chronicle processes both modalities autoregressively within a shared
backbone, so the relative sequence length of each modality directly
influences performance.
For short-series tasks such as TimeCAP, where the time series is
often a single patch, the paired text caption dominates the
mean-pooled representation and leaves temporal features underweighted.
We address this by repeating the TS-token block $r$ times within the
input, appending the same patch embeddings $r$ times without altering
the underlying series, rebalancing the modality ratio without any
architectural change or backbone retraining. Figure~\ref{fig:tok_repeat} sweeps $r \in [1, 128]$ on the three
TimeCAP domains. Averaged across domains, accuracy improves from
$0.681$ at $r{=}1$ to $0.711$ at $r{=}64$, AUC from $0.760$ to
$0.790$, and macro-F1 from $0.602$ to $0.642$, before degrading at
$r \geq 96$ as attention is diluted across many identical copies.
Weather, which has the longest natural TS context, shows the largest
gain; Finance and Healthcare are largely flat, limited by
class-imbalance ceilings rather than representational quality.
The main-paper MM-CLS result (Table~\ref{tab:mm-cls-main}) uses
$r{=}1$ and tuned setting $r{=}64$ to compare to MM-TSFlib and FM Fusion baselines; this ablation
shows roughly 3 accuracy and 4 AUC points of headroom with tuned $r$.

\section{Limitations and Future Work}
\label{sec:limitations}

Several limitations of the current work suggest directions for future research.
First, the forecasting gap to dedicated TSFMs reflects a compute trade-off: our text-heavy 92/8 mix was chosen to keep language understanding competitive with scale-matched text-only models, and closing this gap likely requires more compute or a curriculum strategy (e.g., TS-only pretraining followed by joint continued pretraining) rather than a different architecture.
Second, our causal next-patch objective unifies the text and TS streams but compounds errors over long horizons; hybrid schemes that retain causal attention for text while applying bidirectional attention and contiguous patch masking for time series~\citep{wen2026patchtst-fm} could substantially improve long-horizon zero-shot forecasting within our architecture.
Third, Stage 2 introduces explicit cross-modal supervision for only $5\%$ of TS tokens, yet the consistent zero-shot to linear-probe gain on Time-MMD indicates that substantial cross-modal information remains latent in the frozen backbone; a larger interleaved alignment stage with millions of paired examples is the single most promising direction for improving it.
Finally, we targeted a frozen representation backbone rather than a conversational agent, leaving open whether \modelname{} can serve as a retrieval encoder for time series or, after instruction tuning, as the basis for temporal reasoning in the spirit of ChatTS~\citep{xie2024chatts} or MSE-ITT~\citep{koval2025msitt}.

\section{Conclusion}
\label{sec:conclusion}

We presented \modelname{}, a 324M-parameter decoder-only transformer trained from scratch on natural language and time series within a single shared backbone. Across five benchmarks—NLU, GIFT-Eval, UCR/UEA, Time-MMD, and TimeCAP—\modelname{} matches scale-matched LLMs on language understanding, sets a new bar for frozen-embedding time series classification, and outperforms every supervised fusion baseline on multimodal forecasting, demonstrating that text and time series can share a transformer backbone without catastrophic interference. Our results challenge the prevailing assumption that multimodal time series models must adapt a pretrained LLM, and suggest that joint pretraining from scratch is a more direct path to general-purpose temporal representations. The remaining forecasting gap to dedicated TSFMs is attributable to compute allocation and autoregressive inference, both addressable with scaling and objective refinements; the architecture itself supports both modalities cleanly.

\bibliography{references}

\newpage
\appendix

\section{Model and Code}
All model checkpoints and evaluation code are publicly available at the links below.
The Chronicle Stage~1 and Stage~2 checkpoints are hosted on Hugging Face at \url{[HUGGINGFACE_LINK]}.
Evaluation code is available at \url{[GITHUB_LINK]}.

\section{Full Implementation Details}
\label{app:impl}

\paragraph{Architectural summary.}
The model is a 16-layer decoder-only transformer with $d{=}1024$,
8 GQA query heads, 4 KV heads (head dim 128), and SwiGLU
MLPs~\citep{shazeer2020gluvariantsimprovetransformer} of hidden dimension
$\lceil 8d/3 \rceil$ rounded up to the nearest multiple of 256. Each
block applies pre-norm RMSNorm before the attention and MLP submodules
and uses standard residual connections; no additional residual-stream
modifications are introduced. RoPE~\citep{su2024rope} positional
encodings use base frequency $5\!\times\!10^5$. QK normalization is
applied within attention before the dot product. We use
FlashAttention~\citep{dao2022flashattention} for efficient causal
self-attention; KV-caching support is included for autoregressive
inference. Logit soft-capping via $15 \cdot \tanh(\ell/15)$ is applied to text outputs.

\paragraph{Input/output interfaces.}
Text embeddings are produced by a single learned table
($|\mathcal{V}|{=}131{,}072$, dimension $d$) and tied with the language
model output head. Time series patch embeddings are produced by a
single bias-free linear layer
$\mathbf{W}_p \in \mathbb{R}^{d \times 4P}$ followed by RMSNorm,
applied to the $4P$-dimensional patch features
$[\mathbf{r}; \mathbf{v}; \mathbf{m}; \mathbf{c}]$ described in
Section~\ref{sec:ts_repr}. The quantile head consists of an RMSNorm
followed by a bias-free linear projection
$\mathbb{R}^d \to \mathbb{R}^{P \cdot Q}$ with $Q{=}21$.

\paragraph{Tokenizer.}
The text tokenizer is a byte-level BPE vocabulary of $131{,}072$
tokens trained from scratch with RustBPE on a $50$B-character training corpus. The training mixture consists of $62.5\%$
FineWeb-Edu~\citep{penedo2024fineweb} and $37.5\%$
Dolmino-mix-1124~\citep{olmo2024dolmino} (\texttt{dolmino\_ratio}=$0.375$),
matching the proportions used during model pretraining.

\paragraph{Optimizer groups.}
Parameters are partitioned into a Muon group and three AdamW groups.
Muon~\citep{jordan2024muon} (Newton--Schulz, 5 steps, momentum 0.95) is
applied to all 2D weight matrices in the transformer blocks (attention
$Q$, $K$, $V$, output projection, and SwiGLU $w_1$, $w_2$, $w_3$) at
LR $= 0.02$. AdamW is used for: the token embedding table at LR $= 0.2$;
the (untied) lm\_head at LR $= 0.004$ when applicable; and the patch
projection, quantile head, and all RMSNorm scales (including the
post-embedding norm, the final norm, and the per-block pre-norms) at
LR $= 0.002$. AdamW uses $\beta_1{=}0.8$, $\beta_2{=}0.95$,
$\epsilon{=}10^{-10}$, and weight decay $0$. All AdamW learning rates
are scaled by $\sqrt{768/d}$ to preserve update magnitudes across
model dimensions.

\paragraph{Stage 1: schedule and batching.}
The learning rate follows a three-phase schedule: a 40-step linear
warmup, a constant phase, and linear decay over the final 65\% of
training. We train for 47{,}683 steps at sequence length $2048$ with a
device micro-batch of 48 and 16 gradient accumulation steps, giving a
global batch size of $3{,}145{,}728$ tokens, yielding ${\sim}150$B
total tokens (${\sim}138$B text, ${\sim}12$B time series patches).
At each step, the batch composition (text-only vs.\ time-series-only)
is sampled and broadcast across all data-parallel ranks before any
forward computation, so the gradient at every step is computed
exclusively over a single modality.

\paragraph{Stage 2: extended context and multimodal alignment.}
Stage 2 reloads the stage-1 checkpoint, extends sequence length to
$4096$, and continues training. Within the time-series stream, $5\%$
of tokens are drawn from interleaved alignment data: the alignment
subset of ChatTS~\citep{xie2024chatts} (synthetic series paired with
descriptive text labels) and the time series description corpus of
\citet{merrill2024language} (natural-language descriptions of temporal
patterns). The remaining $95\%$ of TS tokens use the stage-1 unimodal
corpus. For multimodal alignment batches, the loss combines
$\mathcal{L}_{\mathrm{CE}}$ at text positions and
$\mathcal{L}_{\mathrm{QL}}$ at TS positions, with the same global
weighting ($w_{\mathrm{text}}{=}1.0$, $w_{\mathrm{TS}}{=}2.5$). The
text/TS micro-batch ratio remains $0.92/0.08$, the optimizer state is
preserved, and the learning rate continues its decay schedule from the
end of stage 1.

\paragraph{Pretraining data.}
Text comes from a mixture of FineWeb-Edu~\citep{penedo2024fineweb} and
the Dolmino-mix-1124 sub-mixture~\citep{olmo2024dolmino} (DCLM, FLAN,
math, peS2o, Wikipedia, StackExchange) with the dolmino mix ratio set
to $0.334$ during pretraining. Time series data comes from
GiftEvalPretrain (${\sim}900$GB), augmented online with KernelSynth
(2--5 kernels from 33 generators) and per-batch jitter, scaling, and
mixup (Appendix~\ref{app:synthetic}). We do \emph{not} include explicit
multimodal batches during stage 1. Stage 2 introduces the
small alignment slice described above. For pretraining we use two H100 80GB GPUs. Total training for both stages takes roughly one week. 

\paragraph{Weight initialization.}
Linear layers in the transformer use a fan-scaled normal init,
$\mathcal{N}(0,\, \sigma)$ with
$\sigma = \min(1, \sqrt{\mathrm{fan\_out}/\mathrm{fan\_in}})/\sqrt{\mathrm{fan\_in}}$.
The token embedding table is initialized with $\mathcal{N}(0, 0.02)$
when weight tying is enabled. RMSNorm scales are initialized to one.
The output projections of attention and the SwiGLU $w_3$, as well as
the (untied) lm\_head when present, are zero-initialized to keep the
residual stream near identity at initialization. The patch projection
is initialized with the standard fan-scaled normal; the quantile head's
linear projection is initialized to zero.

\section{Downstream Evaluation Setup}
\label{app:eval_setup}

This appendix consolidates the downstream evaluation protocol for
TS classification (UCR), multimodal classification (TimeCAP), and
multimodal forecasting (Time-MMD). All settings here apply to every
method in the corresponding tables, baselines and \modelname{} alike,
unless explicitly noted otherwise.

\subsection{Common Settings}

\begin{table}[h]
\centering
\caption{Settings shared across all downstream evaluations.}
\label{tab:common_settings}
\small
\setlength{\tabcolsep}{6pt}
\begin{tabular}{l l}
\toprule
\textbf{Setting} & \textbf{Value} \\
\midrule
Seed                                 & $1337$ \\
TS normalization                     & instance z-score \\
\modelname{} patch length            & $32$ \\
TimeCAP split                        & stratified $70/10/20$ train/val/test \\
Time-MMD forecasting split            & chronological $70/10/20$, no shuffle \\
UCR TS-CLS split                     & aeon default train/test split \\
TimeCAP MM-CLS max text length       & $384$ tokens \\
TimeCAP MM-CLS trainable loss        & class-balanced cross-entropy \\
\bottomrule
\end{tabular}
\end{table}

The TimeCAP class-balanced cross-entropy weights each class by the
inverse of its training-set frequency, normalized to sum to the number
of classes. This is applied uniformly to every trainable head in the
MM classification table (MM-TSFlib supervised fusion, FM Fusion with
frozen encoders, and the \modelname{} head) so that no method gains an
artificial advantage from majority-class collapse.

\subsection{TimeCAP Multimodal Classification}
\label{app:eval_timecap}

TimeCAP~\citep{gao2024timecap} pairs short multivariate time series
with GPT-4-generated text summaries. We evaluate on three reporting
domains. Weather labels are collapsed from the original
city-specific labels into binary rain / no-rain. Healthcare is reported
as the mean of two underlying binary tasks (in-hospital mortality and
disease test-positive prediction); each is evaluated separately under
the same protocol and the per-domain numbers in
Table~\ref{tab:mm-cls-full} are their average. Dataset
statistics for the four underlying classification tasks are summarized
in Table~\ref{tab:timecap_stats}.

\begin{table}[h]
\centering
\caption{TimeCAP MM classification dataset statistics.
  ``TS shape'' is (steps, channels) for multivariate series and
  (steps,) for univariate. Caption length is reported in whitespace-
  delimited word counts.}
\label{tab:timecap_stats}
\small
\setlength{\tabcolsep}{4pt}
\resizebox{\textwidth}{!}{%
\begin{tabular}{l r r l l r}
\toprule
\textbf{Task} & \textbf{Samples}
  & \textbf{Train\,/\,Val\,/\,Test}
  & \textbf{Classes (counts)}
  & \textbf{TS shape}
  & \textbf{Caption words (mean / max)} \\
\midrule
Weather               & $5{,}652$ & $3955\,/\,566\,/\,1131$ & no rain ($4149$), rain ($1503$) & $(24, 5)$ & $132.6\,/\,196$ \\
Finance               & $1{,}238$ & $866\,/\,124\,/\,248$    & class 1 ($857$), 2 ($211$), 0 ($170$) & $(9,)$ & $160.6\,/\,228$ \\
Healthcare mortality  & $375$     & $262\,/\,38\,/\,75$      & False ($260$), True ($115$) & $(4,)$ & $153.8\,/\,212$ \\
Healthcare positive   & $427$     & $298\,/\,43\,/\,86$      & False ($294$), True ($133$) & $(6,)$ & $154.2\,/\,199$ \\
\bottomrule
\end{tabular}}
\end{table}

\paragraph{Baselines.}
MM-TSFlib fusion baselines pair a trainable time series encoder
(DLinear, PatchTST, or TimesNet) with a frozen pretrained text encoder
(BERT or GPT-2) and a trainable two-layer MLP fusion head. The TS
encoder and head are trained end-to-end for $100$ epochs at learning
rate $10^{-3}$ with batch size $8$. FM Fusion baselines replace the
trainable TS encoder with a frozen pretrained TS foundation model
(Chronos-2, Moirai-2, or TimesFM) and train only the fusion head under
the same schedule. All baselines use the corrected TimeCAP labels,
class-balanced cross-entropy, and a maximum text length of $384$
tokens.

\paragraph{\modelname{}.}
For the linear-probe setting, we feed the joint text--time-series input to a
single frozen \modelname{} backbone and train only a two-layer MLP classification
head. The head uses the same optimizer settings as the fusion baselines:
$100$ epochs, learning rate $10^{-3}$, batch size $8$, dropout $0.1$,
class-balanced cross-entropy, and \texttt{mean\_full} pooling over the backbone
outputs. Because \modelname{} is causally autoregressive, short TimeCAP series
can be underrepresented relative to the accompanying text. We therefore repeat
the TS-token block $r$ times within the input, without changing the underlying
time series or updating the backbone, and sweep
$r \in \{1,2,4,8,16,32,48,64,96,128\}$ in Section~\ref{sec:tok_repeat}.
Macro-F1 peaks at $r{=}32$, while average AUC peaks at $r{=}64$ ($0.792$ versus
$0.788$ at $r{=}32$); we therefore report both the fair-comparison setting
$r{=}1$ and the tuned setting $r{=}64$ in the main results.

For LoRA experiments, the pretrained backbone weights remain fixed and we train
only the LoRA adapters together with the classification head. We report these
rows separately from the linear-probe results to distinguish frozen-backbone
evaluation from parameter-efficient adaptation.

\subsection{Time-MMD Multimodal Forecasting}
\label{app:eval_timemmd}

Time-MMD~\citep{liu2024timemmd} pairs each of nine
domain-specific multivariate time series with aligned textual fact
reports. Each domain is a single chronological sequence; we use the
MM-TSFlib chronological $70/10/20$ split with frequency-specific
context and horizon settings (Table~\ref{tab:timemmd_stats}).

\begin{table}[h]
\centering
\caption{Time-MMD MM forecasting dataset statistics.
  ``Test windows by horizon'' lists the number of evaluation windows
  produced by each horizon length in the same order as the
  ``Horizons'' column.}
\label{tab:timemmd_stats}
\scriptsize
\setlength{\tabcolsep}{3pt}
\resizebox{\textwidth}{!}{%
\begin{tabular}{l l r r r r l l}
\toprule
\textbf{Domain} & \textbf{Freq.} & \textbf{Rows} & \textbf{Channels}
  & \textbf{Train\,/\,Val\,/\,Test rows}
  & \textbf{Context}
  & \textbf{Horizons}
  & \textbf{Test windows by horizon} \\
\midrule
Agriculture   & monthly & $496$    & $3$ & $347\,/\,50\,/\,99$       & $8$  & $6,8,10,12$    & $94,92,90,88$ \\
Climate       & monthly & $496$    & $2$ & $347\,/\,50\,/\,99$       & $8$  & $6,8,10,12$    & $94,92,90,88$ \\
Economy       & monthly & $423$    & $3$ & $296\,/\,43\,/\,84$       & $8$  & $6,8,10,12$    & $79,77,75,73$ \\
Energy        & weekly  & $1{,}479$ & $9$ & $1035\,/\,149\,/\,295$   & $36$ & $12,24,36,48$  & $284,272,260,248$ \\
Environment   & daily   & $15{,}979$ & $2$ & $11185\,/\,1599\,/\,3195$ & $96$ & $48,96,192,336$ & $3148,3100,3004,2860$ \\
Public Health & weekly  & $1{,}389$ & $8$ & $972\,/\,140\,/\,277$    & $36$ & $12,24,36,48$  & $266,254,242,230$ \\
Security      & monthly & $297$    & $1$ & $207\,/\,31\,/\,59$       & $8$  & $6,8,10,12$    & $54,52,50,48$ \\
Social Good   & monthly & $900$    & $1$ & $630\,/\,90\,/\,180$      & $8$  & $6,8,10,12$    & $175,173,171,169$ \\
Traffic       & monthly & $531$    & $1$ & $371\,/\,54\,/\,106$      & $8$  & $6,8,10,12$    & $101,99,97,95$ \\
\bottomrule
\end{tabular}}
\end{table}

\paragraph{Baselines.}
MM-TSFlib forecasting baselines train a trainable TS encoder
(DLinear, PatchTST, or TimesNet) jointly with a frozen text encoder
(BERT or GPT-2) and a trainable forecasting head. FM Fusion baselines
substitute a frozen pretrained TS foundation model (Chronos-2,
Moirai-2, or TimesFM) for the trainable encoder and train only the
fusion head. All baselines minimize MSE on the Time-MMD training split
and report MAE on the held-out test windows; per-domain numbers in
Table~\ref{tab:mm_forecasting} are averaged across the four horizon
lengths in Table~\ref{tab:timemmd_stats}.

\paragraph{\modelname{}.}
We report two variants. \emph{ZS} is autoregressive next-patch
forecasting, with no head training; predictions are denormalized via the inverse of the
patch-level standardization in Eq.~\ref{eq:norm}. \emph{FT} adds a
forecasting head on top of the frozen backbone with joint text and TS
input; only the head's parameters are updated, using MSE loss with
the same horizon and split settings as the baselines.

\subsection{UCR Time Series Classification}

The 14 UCR datasets used in the main paper are GunPoint, Coffee,
ECG200, FaceFour, OSULeaf, SwedishLeaf, SyntheticControl, Trace,
TwoPatterns, Wafer, Earthquakes, ShapeletSim, Chinatown, and
ItalyPowerDemand, with the official aeon train/test splits in all
cases. Supervised DL baselines (Autoformer, DLinear, FEDformer,
Informer, iTransformer, PatchTST, TimesNet) are trained per-dataset
for $30$ epochs at lr $10^{-3}$ with batch size $16$. TS foundation
model baselines (Chronos-2, Moirai-2, TimesFM) and \modelname{} are
evaluated with a learned linear probe on frozen embeddings under the
same aeon split; we train for $200$ epochs at lr
$10^{-2}$ with weight decay $0$ and batch size $64$ on patch-$32$,
joint multivariate, channel-aware, instance z-scored embeddings, while the
foundation-model probes use the published linear-probe protocol from each model's reference implementation.

\section{Extended Related Work}
\label{app:related_full}
 
This appendix provides a comprehensive discussion of the three research threads that \modelname{} builds upon and extends.
 
\subsection{Time Series Foundation Models}
\label{app:related_tsfm}
 
Foundation models for time series aim to generalize zero-shot across domains and frequencies, analogous to how language models generalize across tasks~\citep{liang2024foundation}.
The field has coalesced around two main input representations.
 
\paragraph{Scalar tokenization.}
Chronos~\citep{ansari2024chronos} tokenizes real-valued series via scaling and quantization and trains a T5-family encoder--decoder with cross-entropy loss.
LLMTIME~\citep{gruver2023llmtime} and Chat-TS take scalar tokenization to its extreme by representing values as digit strings and querying frozen LLMs, demonstrating useful numerical priors at the cost of verbosity and computational overhead.
 
\paragraph{Patch-based tokenization.}
A complementary line represents series as \emph{patches}, contiguous windows projected to dense embeddings.
PatchTST~\citep{nie2023patchtst} introduced patch-based tokenization for supervised forecasting, demonstrating that ``a time series is worth 64 words.''
TimesFM~\citep{das2024timesfm} scaled a decoder-only patch transformer to 200M parameters with pretraining on large-scale corpora.
PatchTST-FM~\citep{wen2026patchtst-fm} revisited the generic transformer as a foundation model baseline, adding gated residual projections, a 99-quantile output head, and cumulative patch masking (CPM), achieving state-of-the-art on GIFT-Eval at 260M parameters.
Moirai~\citep{woo2024moirai} addressed heterogeneous frequencies with frequency-specific projections within a masked encoder;
Moirai-2~\citep{liu2026moirai20timeseries} extended this with improved architectures and training.
MOMENT~\citep{goswami2024moment} trains a backbone with lightweight task-specific decoders for multiple tasks simultaneously, while UniTS~\citep{gao2024units} pursues multi-task generalization via unified token representations.
TiRex~\citep{auer2025tirex}, Toto~\citep{cohen2025toto}, and YingLong~\citep{wang2025yinglong} represent further entries on the GIFT-Eval leaderboard.
 
Our architecture draws on PatchTST-FM and TimesFM (patch-based, decoder-only, quantile output) but differs fundamentally in being trained jointly with natural language from scratch.
We compare against these models on GIFT-Eval (Section~\ref{sec:eval_pretrain_gift}), where published leaderboard scores provide a direct zero-shot comparison, and on UCR classification (Section~\ref{sec:eval_tsclass}), where Chronos-2 and Moirai-2 frozen embeddings serve as foundation model baselines.
 
\subsection{Multimodal Text and Time Series Models}
\label{app:related_mm}
 
A rapidly growing body of work connects language models to time series.
We organize these approaches by their architectural paradigm and highlight their evaluation limitations.
 
\paragraph{Frozen LLM approaches.}
LLMTIME~\citep{gruver2023llmtime} queries frozen GPT-3/LLaMA with digit-string representations of time series, demonstrating zero-shot forecasting capability but inheriting the full computational cost of large language models and producing no learnable temporal representations.
GPT4MTS~\citep{jia2024gpt4mts} constructs multimodal prompts combining textual context with numerical time series data and feeds them to frozen LLMs.
\citet{zhou2023gpt4ts} showed that frozen LLMs, fine-tuned only at input/output projections, yield competitive forecasting performance.
However, these approaches treat the language model as a black box; the temporal representations are constrained to the text embedding space, which was never designed for continuous numerical data.
 
\paragraph{Adapted LLM approaches.}
Time-LLM~\citep{jin2024timellm} reprograms patch embeddings into text prototypes with a frozen LLM backbone.
GPT4TS~\citep{zhou2023gpt4ts} fine-tunes only the normalization layers of GPT-2.
ChatTS~\citep{xie2024chatts} encodes time series patches through a shallow MLP and concatenates them with text embeddings before feeding a fine-tuned Qwen2.5-14B backbone, using synthetic QA pairs (TSEvol) to address data scarcity.
ChatTime~\citep{wang2024chattime} instruction-fine-tunes a decoder-only LLM for bidirectional text and time series generation, achieving 99.9\% of Chronos's zero-shot accuracy with only 4\% of the pretraining data by leveraging the pretrained LLM's existing representations.
MoAT~\citep{lee2024moat} introduces a two-stage framework: first optimizing forecasts from decomposed time series and text embeddings, then fusing via an offline MLP synthesis.
TaTs~\citep{li2025tats} treats text embeddings as auxiliary time series variables, capturing what the authors call ``chronological textual resonance,'' periodic patterns in text representations that mirror the numerical series.
MSE-ITT~\citep{koval2025msitt} extends LLaMA-3-8B with modality-specific expert layers for financial forecasting from interleaved text and time series.
Time-VLM~\citep{zhong2025timevlmexploringmultimodalvisionlanguage} bridges temporal, visual, and textual modalities using frozen vision-language models.
 
All of these approaches share a fundamental limitation: \emph{they start from a pretrained language model}, meaning the backbone's representations were shaped entirely by text before any exposure to temporal data.
The time series modality must adapt to a representational space that was not designed for it, and the resulting models inherit the language model's parameter count, vocabulary, and computational requirements, even when the downstream task is purely temporal.
 
\paragraph{Fusion and benchmark approaches.}
Time-MMD~\citep{liu2024timemmd} provides a multi-domain benchmark pairing time series with textual reports across nine domains and introduces \textbf{MM-TSFlib}, a fusion library that has become a standard reference protocol for text-augmented time series. Under MM-TSFlib, a trainable time series encoder (e.g., DLinear, PatchTST, TimesNet) is paired with a frozen pretrained text encoder (BERT or GPT-2) and a trainable MLP fusion head; the TS encoder and head are trained end-to-end on each downstream dataset, while the text encoder remains frozen. Subsequent text-augmented time series studies have adopted MM-TSFlib as a benchmarking baseline; we use it directly as our multimodal fusion comparison in Section~\ref{sec:eval_downstream}.
TimeCAP~\citep{gao2024timecap} uses LLM agents to generate contextual descriptions and combines predictions from a multimodal predictor with a pretrained LLM. Recent surveys on multimodal time series~\citep{liu2025crossmodalitymodelingtimeseries} provide comprehensive taxonomies of fusion strategies.
\citet{zhang2025whenmm} systematically investigate when multimodal integration yields gains, finding that benefits are ``highly condition-dependent'' and ``neither universal nor always aligned with intuition.''
 
\paragraph{Critical evaluation gap.}
A striking pattern across this literature is the narrowness of evaluation.
ChatTime compares against Chronos and GPT4TS for forecasting but does not evaluate language understanding.
ChatTS evaluates time series understanding but not against GIFT-Eval, UCR, or NLU benchmarks.
MoAT, TaTs, and GPT4MTS evaluate only multimodal forecasting on their own datasets.
MSE-ITT compares against multimodal and financial baselines but does not benchmark against dedicated TSFMs on standard time series tasks.
\emph{No prior multimodal text and time series model has been evaluated against both dedicated TSFMs on time series benchmarks and dedicated LLMs on language understanding benchmarks.}
This creates a fundamental ambiguity: when a multimodal model reports improved forecasting, it is unclear whether the improvement stems from genuine cross-modal learning or simply from the text providing complementary information that a strong TSFM baseline would render unnecessary.
Our evaluation protocol addresses this gap directly by testing \modelname{} against the best models in \emph{each} modality on \emph{their own} benchmarks.
 
\paragraph{Negative results on LLMs for time series.}
Several recent works have questioned the value of language model priors for temporal tasks.
\citet{tan2024llmts} ablated three top-tier LLM-for-TS methods and found that LLMs ``fail to convincingly improve time series forecasting'' while ``significantly increasing computational costs.''
Re-initializing LLM weights prior to forecasting had no impact on performance, suggesting that pretrained language representations do not transfer to temporal modeling.
\citet{merrill2024language} found that LLMs struggle to reason about time series encoded as text, motivating modality-native representations.
These findings suggest that simply bolting time series onto a language model, the approach taken by all prior multimodal work, is fundamentally limited.
\modelname{} takes a different path: rather than adapting a language model for time series, we train a single model for both from scratch, allowing the architecture to develop representations suitable for both modalities simultaneously.
 
\subsection{Small Language Models}
\label{app:related_lm}
 
GPT-2~\citep{radford2019gpt2} demonstrated that decoder-only transformers trained with next-token prediction produce capable few-shot learners.
Subsequent models have pushed zero-shot language understanding to strong levels at sub-1B scale:
Qwen2~\citep{yang2024qwen2} at 500M parameters achieves 0.476 average accuracy on our NLU suite;
LLaMA-3.2~\citep{grattafiori2024llama3} at 1.2B achieves 0.531;
Gemma-3-270M-PT~\citep{team2025gemma3} at 270M achieves 0.406;
and LFM-2-350M~\citep{liquid2025lfm2} at 350M achieves 0.449.
These models represent the current frontier of what is achievable with compact transformer architectures trained exclusively on text. They are also typically trained on hundreds of billions to several trillion text tokens, substantially more than our total compute budget allows.
We compare \modelname{} against all five to verify that, under our text-heavy 92/8 mix, devoting approximately 8\% of training compute to time series does not cause catastrophic interference.
The fact that \modelname{} matches Gemma-3-270M-PT despite its dual training objective establishes an important proof point: a shared transformer backbone can accommodate both text and time series without degrading either modality's performance relative to scale-matched specialists, provided the token mix is chosen to keep the language stream competitive.
 
\section{Synthetic Training Data}
\label{app:synthetic}

Online synthetic augmentation is applied to time-series-only batches
during training, controlled by a per-batch probability of $0.20$. Series
are generated on-the-fly in a background worker thread; generation takes
approximately $1$ to $3$\,ms per series at length 32k, introducing no
data-loading bottleneck.

A bank of 33 kernel generators is defined at module load time, spanning
smooth trends, periodic patterns, stochastic processes, discrete
waveforms, and noise models (Table~\ref{tab:kernel_bank}). For each
synthetic series, 2--5 kernels are sampled without replacement and
combined via one of two modes:

\begin{itemize}
  \item \textbf{Additive} (80\%): $x(t) = \sum_i k_i(t)$.
  \item \textbf{Mixed multiplicative} (20\%): kernels are combined
        iteratively; each subsequent kernel is either added or
        multiplied (after shifting to a positive range) with
        probability $0.40$ per kernel.
\end{itemize}

All kernels operate on normalized time
$t_n = \mathrm{linspace}(0,1,L)$ and are vectorized. Duplicate entries
in the bank increase sampling frequency for empirically useful kernels
(RBF short/long, periodic short/long, rational quadratic, damped
oscillation), following the emphasis in Chronos
KernelSynth~\citep{ansari2024chronos}. After composition, Inf values are clipped to $\pm 5$ before combination and $\pm 10^7$ after; output is cast to float32. Beyond KernelSynth, with 50\% probability
per time-series batch we additionally apply jitter (additive Gaussian
noise), scaling (multiplicative perturbation), and intra-batch mixup.
\begin{table}[h]
\centering
\caption{KernelSynth generator bank (33 entries). Duplicates are listed in the ``\# entries'' column and increase sampling weight for empirically useful kernels.}
\label{tab:kernel_bank}
\small
\setlength{\tabcolsep}{5pt}
\resizebox{\columnwidth}{!}{%
\begin{tabular}{l l r}
\toprule
\textbf{Category} & \textbf{Implementation} & \textbf{\# entries} \\
\midrule
RBF smooth & $\frac{1}{R}\sum_r \cos(\omega_r t + \phi_r)$, $\omega_r \sim \mathcal{N}(0, 1/\ell_s)$, $R{=}32$ RFF & 5 \\
Periodic & $A\sin(2\pi t/p + \phi)$, $A \sim \mathrm{Unif}(0.5, 2)$ & 5 \\
Periodic + harmonics & Base + 2 overtones at amplitudes $A/2$, $A/3$ with independent phases & 1 \\
Rational Quadratic & RFF with Gamma-distributed scales: $\omega_r \sim \mathrm{scale} \cdot \mathcal{N}(0,1)/\ell_s$, $\mathrm{scale} \sim \Gamma(\alpha, 1/\alpha)$ & 2 \\
Linear trend & $at + b$, $a \sim \mathrm{Unif}(-3, 3)$ & 1 \\
Polynomial & $\mathrm{polyval}(\mathbf{c}, t_{\mathrm{norm}})$, coefficients $\sim \mathrm{Unif}$ & 2 \\
Log trend & $c \cdot \log(t)$, $c \sim \mathrm{Unif}(-2, 2)$ & 1 \\
Random walk & Cumulative sum of Gaussian steps; drift $\mu \sim \mathrm{Unif}(-0.01, 0.01)$ & 2 \\
Level shifts & 1--3 abrupt shifts at random positions in the middle 80\% & 1 \\
Discrete waves & Period $\in [0.05, 0.40]$, amplitude $\in [0.5, 2.0]$, random phase/offset & 3 \\
Damped oscillation & $Ae^{-\gamma t}\sin(2\pi t/p + \phi)$, $\gamma \sim \mathrm{Unif}(1, 8)$ & 2 \\
White noise & $\mathcal{N}(0, \sigma)$ & 3 \\
Heteroskedastic noise & $\epsilon_t \sim \mathcal{N}(0,\, \sigma \cdot e^{0.5 k(t)})$, envelope modulated by RBF-drawn signal & 1 \\
Periodic noise & $\mathcal{N}(0, 0.3) \cdot (1 + A(\sin(2\pi t/p + \phi) \cdot 0.5 + 0.5))$ & 1 \\
Step function & 3--11 constant-level segments with random transitions & 1 \\
Exponential growth/decay & $e^{rt} - 1$, $r \sim \mathrm{Unif}(-3, 3)$ & 1 \\
Constant & Flat baseline $c \sim \mathrm{Unif}(-2, 2)$ & 1 \\
\midrule
\textbf{Total} & & \textbf{33} \\
\bottomrule
\end{tabular}%
}
\end{table}

\subsection{Full Multimodal Classification Results}
\label{app:mm_cls_full}

Table~\ref{tab:mm-cls-full} reports per-domain accuracy,
macro-F1, and AUC for all methods on TimeCAP. Within each baseline
category, BERT-paired models appear before GPT2-paired models.
Summary averages are reported in Table~\ref{tab:mm-cls-main}.

\begin{table*}[t]
\centering
\small
\setlength{\tabcolsep}{3pt}
\caption{Multimodal classification on TimeCAP by domain. Values are mean $\pm$ standard deviation over 3 seeds (0, 1, 2).}
\label{tab:mm-cls-full}
\resizebox{\textwidth}{!}{%
\begin{tabular}{llcccccccc}
\toprule
\textbf{Cat.} & \textbf{Model} & \multicolumn{2}{c}{\textbf{Weather}} & \multicolumn{2}{c}{\textbf{Finance}} & \multicolumn{2}{c}{\textbf{Healthcare}} & \multicolumn{2}{c}{\textbf{Average}} \\
 & & F1 $\uparrow$ & AUC $\uparrow$ & F1 $\uparrow$ & AUC $\uparrow$ & F1 $\uparrow$ & AUC $\uparrow$ & F1 $\uparrow$ & AUC $\uparrow$ \\
\midrule
MM-TSFlib & DLin+BERT & $0.600 \pm 0.013$ & $0.676 \pm 0.011$ & $0.367 \pm 0.010$ & $0.641 \pm 0.036$ & $0.796 \pm 0.051$ & $0.902 \pm 0.030$ & $0.588 \pm 0.016$ & $0.739 \pm 0.024$ \\
 & DLin+GPT2 & $0.572 \pm 0.009$ & $0.643 \pm 0.017$ & $0.326 \pm 0.057$ & $0.655 \pm 0.018$ & $0.793 \pm 0.024$ & $0.875 \pm 0.036$ & $0.564 \pm 0.026$ & $0.724 \pm 0.017$ \\
 & PTST+BERT & $0.566 \pm 0.017$ & $0.613 \pm 0.021$ & $0.368 \pm 0.052$ & $0.643 \pm 0.024$ & $0.800 \pm 0.004$ & $0.900 \pm 0.024$ & $0.578 \pm 0.022$ & $0.719 \pm 0.022$ \\
 & PTST+GPT2 & $0.461 \pm 0.066$ & $0.556 \pm 0.057$ & $0.363 \pm 0.010$ & $0.682 \pm 0.019$ & $0.793 \pm 0.009$ & $0.882 \pm 0.033$ & $0.539 \pm 0.021$ & $0.707 \pm 0.034$ \\
 & TNet+BERT & $0.597 \pm 0.057$ & $0.712 \pm 0.013$ & $0.364 \pm 0.015$ & $0.635 \pm 0.041$ & $0.808 \pm 0.017$ & $0.902 \pm 0.026$ & $0.589 \pm 0.021$ & $0.750 \pm 0.026$ \\
 & TNet+GPT2 & $0.617 \pm 0.037$ & $0.729 \pm 0.018$ & $0.330 \pm 0.056$ & $0.660 \pm 0.048$ & $0.786 \pm 0.032$ & $0.871 \pm 0.036$ & $0.577 \pm 0.019$ & $0.754 \pm 0.028$ \\
\midrule
FM Fusion & BERT+Chr2 & $0.588 \pm 0.010$ & $0.631 \pm 0.018$ & $0.388 \pm 0.054$ & $0.647 \pm 0.028$ & $0.794 \pm 0.012$ & $0.902 \pm 0.029$ & $0.590 \pm 0.021$ & $0.726 \pm 0.023$ \\
 & BERT+Moi2 & $0.617 \pm 0.022$ & $0.702 \pm 0.023$ & $0.351 \pm 0.041$ & $0.651 \pm 0.026$ & $0.797 \pm 0.019$ & $0.901 \pm 0.033$ & $0.588 \pm 0.004$ & $0.751 \pm 0.025$ \\
 & BERT+TFM & $0.436 \pm 0.021$ & $0.577 \pm 0.030$ & $0.273 \pm 0.000$ & $0.503 \pm 0.102$ & $0.786 \pm 0.009$ & $0.898 \pm 0.030$ & $0.498 \pm 0.006$ & $0.659 \pm 0.023$ \\
 & GPT2+Chr2 & $0.284 \pm 0.121$ & $0.542 \pm 0.031$ & $0.294 \pm 0.037$ & $0.597 \pm 0.143$ & $0.785 \pm 0.023$ & $0.880 \pm 0.035$ & $0.455 \pm 0.043$ & $0.673 \pm 0.056$ \\
 & GPT2+Moi2 & $0.588 \pm 0.015$ & $0.690 \pm 0.022$ & $0.252 \pm 0.150$ & $0.643 \pm 0.029$ & $0.786 \pm 0.037$ & $0.885 \pm 0.024$ & $0.542 \pm 0.038$ & $0.739 \pm 0.015$ \\
 & GPT2+TFM & $0.423 \pm 0.000$ & $0.562 \pm 0.029$ & $0.273 \pm 0.000$ & $0.451 \pm 0.077$ & $0.745 \pm 0.101$ & $0.870 \pm 0.054$ & $0.480 \pm 0.034$ & $0.628 \pm 0.018$ \\
\midrule
\textbf{\modelname{}} & Stage 1 LP (r=1) & $0.564 \pm 0.021$ & $0.641 \pm 0.034$ & $0.426 \pm 0.024$ & $0.683 \pm 0.024$ & $0.790 \pm 0.028$ & $0.874 \pm 0.048$ & $0.593 \pm 0.021$ & $0.733 \pm 0.030$ \\
 & Stage 1 LP (r=64) & $0.608 \pm 0.018$ & $0.705 \pm 0.010$ & $0.423 \pm 0.009$ & $0.662 \pm 0.003$ & $0.793 \pm 0.021$ & $0.867 \pm 0.042$ & $0.608 \pm 0.010$ & $0.745 \pm 0.014$ \\
 & Stage 1 LoRA (r=1) & $0.602 \pm 0.012$ & $0.641 \pm 0.029$ & $0.374 \pm 0.032$ & $0.684 \pm 0.030$ & $0.828 \pm 0.011$ & $0.891 \pm 0.017$ & $0.601 \pm 0.011$ & $0.739 \pm 0.017$ \\
 & Stage 1 LoRA (r=64) & $0.636 \pm 0.006$ & $0.717 \pm 0.018$ & $0.295 \pm 0.038$ & $0.669 \pm 0.005$ & $0.820 \pm 0.042$ & $0.904 \pm 0.039$ & $0.584 \pm 0.024$ & $0.763 \pm 0.011$ \\
 & Stage 2 LP (r=1) & $0.561 \pm 0.039$ & $0.635 \pm 0.040$ & $0.428 \pm 0.011$ & $0.685 \pm 0.014$ & $0.794 \pm 0.042$ & $0.872 \pm 0.048$ & $0.594 \pm 0.025$ & $0.731 \pm 0.029$ \\
 & Stage 2 LP (r=64) & $0.606 \pm 0.013$ & $0.696 \pm 0.008$ & $0.406 \pm 0.014$ & $0.683 \pm 0.010$ & $0.804 \pm 0.025$ & $0.872 \pm 0.038$ & $0.605 \pm 0.011$ & $0.750 \pm 0.014$ \\
 & Stage 2 LoRA (r=1) & $0.599 \pm 0.013$ & $0.644 \pm 0.024$ & $0.365 \pm 0.080$ & $0.656 \pm 0.074$ & $0.821 \pm 0.013$ & $0.892 \pm 0.019$ & $0.595 \pm 0.030$ & $0.731 \pm 0.032$ \\
 & Stage 2 LoRA (r=64) & $0.629 \pm 0.033$ & $0.706 \pm 0.066$ & $0.386 \pm 0.010$ & $0.667 \pm 0.047$ & $0.823 \pm 0.016$ & $0.897 \pm 0.036$ & $0.613 \pm 0.011$ & $0.757 \pm 0.014$ \\
\bottomrule
\end{tabular}}
\end{table*}

\subsection{Effect of Channel-Aware Multivariate Handling}
\label{app:channel_ramp}

To assess the value of preserving channel identity in multivariate time-series inputs, we compare the default joint multivariate representation used by \modelname{}, which includes the channel ramp $\mathbf{c}$ in the patch features, against a mean-channel pooling variant that averages channels before encoding and therefore removes channel identity. All values are for the frozen-backbone linear head.

\begin{figure*}[t]
  \centering
  \includegraphics[width=\textwidth]{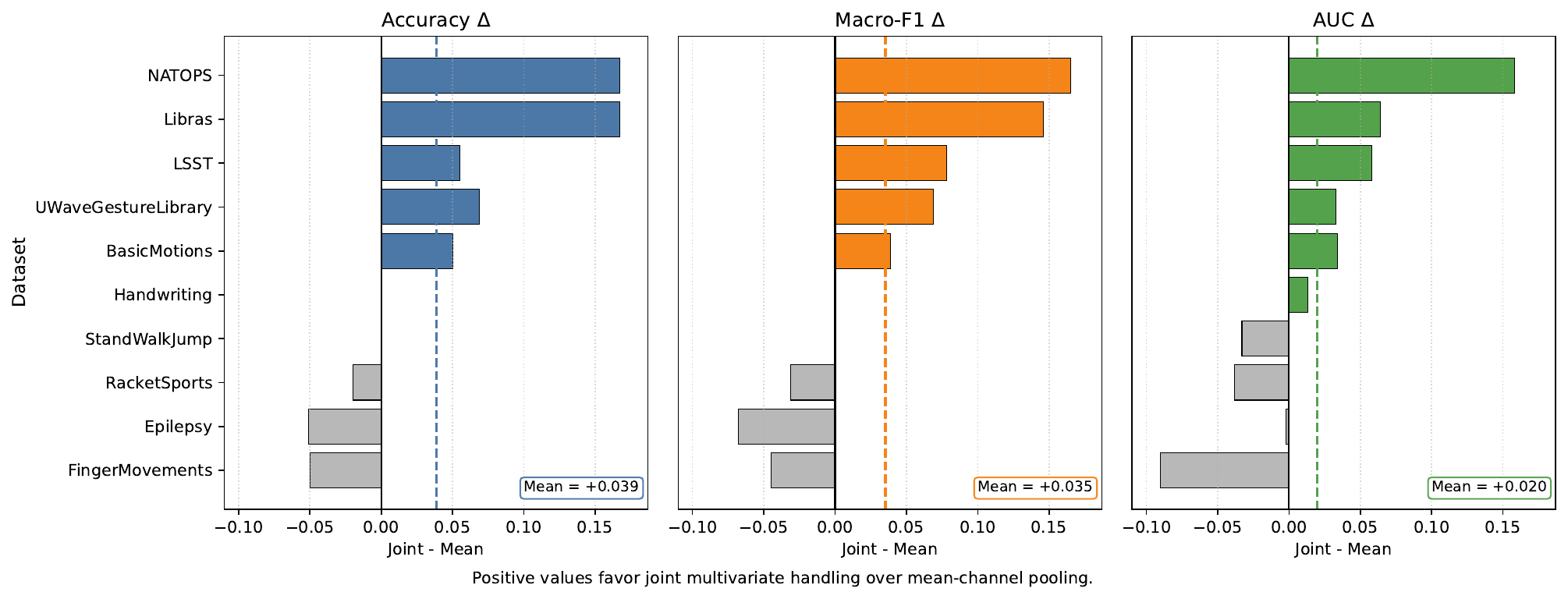}
  \caption{\textbf{Effect of channel-aware multivariate handling on UEA classification.}
  Bars show the per-dataset delta between joint multivariate handling and mean-channel pooling (joint minus mean) for accuracy, macro-F1, and AUC. Positive values favor joint channel-aware handling. Averaged across the 10 multivariate UEA datasets, joint handling improves accuracy by $+0.039$, macro-F1 by $+0.035$, and AUC by $+0.020$.}
  \label{fig:uea-channel-ramp}
\end{figure*}

Figure~\ref{fig:uea-channel-ramp} shows the per-dataset deltas (joint minus mean pooling) on the 10 multivariate UEA datasets used in our evaluation. Averaged across datasets, joint channel-aware handling improves accuracy by $+0.039$, macro-F1 by $+0.035$, and AUC by $+0.020$. The largest gains appear on \textsc{Libras} and \textsc{NATOPS} ($+0.167$ accuracy on both), with additional improvements on \textsc{LSST} and \textsc{UWaveGestureLibrary}. Some datasets favor mean-channel pooling (\textsc{Epilepsy}, \textsc{FingerMovements}, and \textsc{RacketSports}), while \textsc{Handwriting} and \textsc{StandWalkJump} are effectively unchanged in accuracy and F1. Overall, the results indicate that retaining channel identity is beneficial on average for multivariate classification, supporting the use of channel-aware patch features for multivariate inputs.

\section{Per-Dataset UCR/UEA Classification Results}
\label{app:ucr}

\begin{table*}[t]
\centering
\scriptsize
\setlength{\tabcolsep}{2pt}
\caption{Per-dataset UCR/UEA time-series classification results. Values are means over five seeds.}
\label{tab:ucr-uea-full}
\resizebox{\textwidth}{!}{%
\begin{tabular}{llcccccccccccccccccccccccccccccccccccc}
\toprule
Suite & Dataset & \multicolumn{3}{c}{Informer} & \multicolumn{3}{c}{TimesNet} & \multicolumn{3}{c}{Autoformer} & \multicolumn{3}{c}{DLinear} & \multicolumn{3}{c}{iTransformer} & \multicolumn{3}{c}{FEDformer} & \multicolumn{3}{c}{PatchTST} & \multicolumn{3}{c}{Chronos-2} & \multicolumn{3}{c}{TimesFM} & \multicolumn{3}{c}{Moirai-2} & \multicolumn{3}{c}{Chronicle Stage 1} & \multicolumn{3}{c}{Chronicle Stage 2} \\
\cmidrule(lr){3-5} \cmidrule(lr){6-8} \cmidrule(lr){9-11} \cmidrule(lr){12-14} \cmidrule(lr){15-17} \cmidrule(lr){18-20} \cmidrule(lr){21-23} \cmidrule(lr){24-26} \cmidrule(lr){27-29} \cmidrule(lr){30-32} \cmidrule(lr){33-35} \cmidrule(lr){36-38}
 &  & Acc & F1 & AUC & Acc & F1 & AUC & Acc & F1 & AUC & Acc & F1 & AUC & Acc & F1 & AUC & Acc & F1 & AUC & Acc & F1 & AUC & Acc & F1 & AUC & Acc & F1 & AUC & Acc & F1 & AUC & Acc & F1 & AUC & Acc & F1 & AUC \\
\midrule
UCR & GunPoint & 0.576 & 0.573 & 0.634 & 0.580 & 0.539 & 0.603 & 0.605 & 0.599 & 0.716 & 0.757 & 0.756 & 0.840 & 0.708 & 0.701 & 0.842 & 0.656 & 0.650 & 0.779 & 0.653 & 0.648 & 0.711 & 0.528 & 0.397 & 0.548 & 0.712 & 0.682 & 0.862 & 0.931 & 0.931 & 0.981 & 0.919 & 0.919 & 0.965 & 0.851 & 0.850 & 0.936 \\
 & Coffee & 0.536 & 0.349 & 0.397 & 0.536 & 0.349 & 0.588 & 0.536 & 0.349 & 0.525 & 0.900 & 0.871 & 1.000 & 0.593 & 0.441 & 0.962 & 0.536 & 0.349 & 0.563 & 0.536 & 0.349 & 0.673 & 0.536 & 0.410 & 0.555 & 0.571 & 0.426 & 0.987 & 0.964 & 0.964 & 0.995 & 0.893 & 0.892 & 0.988 & 0.893 & 0.870 & 0.996 \\
 & ECG200 & 0.642 & 0.441 & 0.575 & 0.804 & 0.791 & 0.894 & 0.806 & 0.779 & 0.873 & 0.814 & 0.796 & 0.904 & 0.866 & 0.855 & 0.942 & 0.796 & 0.764 & 0.906 & 0.862 & 0.846 & 0.920 & 0.672 & 0.468 & 0.572 & 0.840 & 0.818 & 0.939 & 0.820 & 0.799 & 0.892 & 0.846 & 0.826 & 0.946 & 0.848 & 0.833 & 0.934 \\
 & FaceFour & 0.186 & 0.078 & 0.586 & 0.395 & 0.288 & 0.785 & 0.227 & 0.116 & 0.473 & 0.541 & 0.516 & 0.923 & 0.627 & 0.629 & 0.888 & 0.223 & 0.102 & 0.556 & 0.350 & 0.296 & 0.687 & 0.236 & 0.123 & 0.521 & 0.609 & 0.607 & 0.879 & 0.582 & 0.552 & 0.810 & 0.864 & 0.866 & 0.988 & 0.759 & 0.731 & 0.984 \\
 & OSULeaf & 0.350 & 0.270 & 0.749 & 0.406 & 0.319 & 0.825 & 0.513 & 0.457 & 0.824 & 0.360 & 0.313 & 0.666 & 0.406 & 0.350 & 0.763 & 0.556 & 0.485 & 0.863 & 0.559 & 0.510 & 0.863 & 0.184 & 0.055 & 0.519 & 0.401 & 0.311 & 0.842 & 0.719 & 0.693 & 0.937 & 0.583 & 0.555 & 0.876 & 0.579 & 0.542 & 0.888 \\
 & SwedishLeaf & 0.533 & 0.497 & 0.947 & 0.381 & 0.314 & 0.923 & 0.675 & 0.647 & 0.976 & 0.763 & 0.756 & 0.969 & 0.843 & 0.841 & 0.990 & 0.711 & 0.701 & 0.975 & 0.859 & 0.858 & 0.992 & 0.096 & 0.054 & 0.554 & 0.599 & 0.564 & 0.966 & 0.854 & 0.854 & 0.992 & 0.848 & 0.847 & 0.993 & 0.834 & 0.831 & 0.987 \\
 & SyntheticControl & 0.835 & 0.829 & 0.974 & 0.591 & 0.519 & 0.990 & 0.680 & 0.666 & 0.934 & 0.878 & 0.868 & 0.986 & 0.833 & 0.821 & 0.981 & 0.992 & 0.992 & 0.999 & 0.926 & 0.926 & 0.993 & 0.239 & 0.139 & 0.567 & 0.903 & 0.904 & 0.989 & 0.860 & 0.860 & 0.983 & 0.727 & 0.728 & 0.930 & 0.727 & 0.727 & 0.931 \\
 & Trace & 0.746 & 0.703 & 0.938 & 0.506 & 0.420 & 0.931 & 0.910 & 0.903 & 0.975 & 0.484 & 0.444 & 0.830 & 0.488 & 0.385 & 0.837 & 1.000 & 1.000 & 1.000 & 0.878 & 0.868 & 0.984 & 0.288 & 0.189 & 0.581 & 0.630 & 0.573 & 0.891 & 0.802 & 0.804 & 0.951 & 0.936 & 0.932 & 0.999 & 0.848 & 0.827 & 0.988 \\
 & TwoPatterns & 0.355 & 0.253 & 0.704 & 0.999 & 0.999 & 1.000 & 0.357 & 0.304 & 0.617 & 0.850 & 0.850 & 0.975 & 0.820 & 0.819 & 0.960 & 0.932 & 0.932 & 0.992 & 0.826 & 0.825 & 0.957 & 0.281 & 0.153 & 0.525 & 0.655 & 0.627 & 0.916 & 0.634 & 0.632 & 0.863 & 0.778 & 0.777 & 0.948 & 0.790 & 0.787 & 0.945 \\
 & Wafer & 0.982 & 0.950 & 0.982 & 0.951 & 0.863 & 0.933 & 0.988 & 0.969 & 0.998 & 0.943 & 0.834 & 0.853 & 0.995 & 0.987 & 0.998 & 0.992 & 0.979 & 0.998 & 0.974 & 0.931 & 0.984 & 0.894 & 0.488 & 0.537 & 0.941 & 0.813 & 0.919 & 0.996 & 0.990 & 1.000 & 0.991 & 0.977 & 0.999 & 0.987 & 0.966 & 0.998 \\
 & Earthquakes & 0.748 & 0.428 & 0.680 & 0.748 & 0.428 & 0.666 & 0.748 & 0.428 & 0.681 & 0.630 & 0.545 & 0.583 & 0.732 & 0.471 & 0.500 & 0.748 & 0.428 & 0.688 & 0.738 & 0.440 & 0.619 & 0.748 & 0.428 & 0.506 & 0.581 & 0.436 & 0.507 & 0.722 & 0.468 & 0.656 & 0.760 & 0.536 & 0.645 & 0.774 & 0.555 & 0.682 \\
 & ShapeletSim & 0.499 & 0.333 & 0.536 & 0.779 & 0.737 & 0.947 & 0.559 & 0.482 & 0.662 & 0.526 & 0.523 & 0.521 & 0.491 & 0.489 & 0.496 & 0.656 & 0.583 & 0.972 & 0.519 & 0.445 & 0.546 & 0.496 & 0.357 & 0.502 & 0.709 & 0.706 & 0.757 & 0.789 & 0.788 & 0.856 & 0.603 & 0.603 & 0.623 & 0.608 & 0.605 & 0.682 \\
 & Chinatown & 0.574 & 0.388 & 0.646 & 0.412 & 0.363 & 0.985 & 0.706 & 0.671 & 0.881 & 0.908 & 0.876 & 0.987 & 0.969 & 0.962 & 0.994 & 0.845 & 0.835 & 0.988 & 0.904 & 0.880 & 0.959 & 0.684 & 0.487 & 0.599 & 0.930 & 0.913 & 0.976 & 0.983 & 0.978 & 0.996 & 0.977 & 0.971 & 0.996 & 0.960 & 0.953 & 0.995 \\
 & ItalyPowerDemand & 0.557 & 0.542 & 0.717 & 0.829 & 0.819 & 0.983 & 0.741 & 0.740 & 0.815 & 0.952 & 0.952 & 0.990 & 0.969 & 0.969 & 0.993 & 0.848 & 0.848 & 0.913 & 0.970 & 0.970 & 0.988 & 0.591 & 0.458 & 0.598 & 0.927 & 0.926 & 0.976 & 0.946 & 0.946 & 0.991 & 0.953 & 0.953 & 0.991 & 0.955 & 0.955 & 0.992 \\
\midrule
UEA & BasicMotions & 0.980 & 0.980 & 1.000 & 0.980 & 0.980 & 1.000 & 0.995 & 0.995 & 1.000 & 0.390 & 0.372 & 0.666 & 0.450 & 0.415 & 0.725 & 1.000 & 1.000 & 1.000 & 0.380 & 0.327 & 0.626 & 0.290 & 0.174 & 0.543 & 0.765 & 0.762 & 0.938 & 0.960 & 0.960 & 0.998 & 0.925 & 0.924 & 0.995 & 0.945 & 0.946 & 0.990 \\
 & Epilepsy & 0.478 & 0.481 & 0.701 & 0.848 & 0.849 & 0.977 & 0.642 & 0.645 & 0.871 & 0.370 & 0.358 & 0.575 & 0.309 & 0.297 & 0.532 & 0.922 & 0.920 & 0.987 & 0.877 & 0.875 & 0.975 & 0.272 & 0.138 & 0.502 & 0.897 & 0.894 & 0.989 & 0.943 & 0.945 & 0.996 & 0.945 & 0.943 & 0.992 & 0.939 & 0.934 & 0.987 \\
 & NATOPS & 0.716 & 0.700 & 0.953 & 0.881 & 0.878 & 0.982 & 0.808 & 0.803 & 0.970 & 0.716 & 0.714 & 0.933 & 0.637 & 0.629 & 0.900 & 0.896 & 0.894 & 0.988 & 0.693 & 0.682 & 0.918 & 0.220 & 0.115 & 0.551 & 0.511 & 0.492 & 0.823 & 0.513 & 0.503 & 0.827 & 0.544 & 0.535 & 0.839 & 0.523 & 0.512 & 0.826 \\
 & RacketSports & 0.728 & 0.739 & 0.912 & 0.814 & 0.829 & 0.944 & 0.812 & 0.824 & 0.941 & 0.687 & 0.697 & 0.884 & 0.608 & 0.615 & 0.845 & 0.851 & 0.864 & 0.948 & 0.593 & 0.597 & 0.850 & 0.322 & 0.183 & 0.555 & 0.668 & 0.659 & 0.880 & 0.693 & 0.697 & 0.894 & 0.659 & 0.662 & 0.894 & 0.701 & 0.711 & 0.890 \\
 & UWaveGestureLibrary & 0.410 & 0.397 & 0.828 & 0.606 & 0.608 & 0.907 & 0.520 & 0.507 & 0.887 & 0.801 & 0.792 & 0.964 & 0.676 & 0.668 & 0.925 & 0.652 & 0.643 & 0.914 & 0.785 & 0.783 & 0.944 & 0.139 & 0.053 & 0.525 & 0.221 & 0.132 & 0.740 & 0.500 & 0.508 & 0.863 & 0.627 & 0.615 & 0.921 & 0.636 & 0.622 & 0.919 \\
 & Handwriting & 0.132 & 0.064 & 0.692 & 0.147 & 0.079 & 0.774 & 0.147 & 0.078 & 0.685 & 0.132 & 0.097 & 0.613 & 0.144 & 0.106 & 0.614 & 0.343 & 0.271 & 0.883 & 0.093 & 0.066 & 0.595 & 0.041 & 0.009 & 0.510 & 0.107 & 0.087 & 0.626 & 0.127 & 0.115 & 0.648 & 0.172 & 0.148 & 0.661 & 0.146 & 0.116 & 0.659 \\
 & Libras & 0.569 & 0.559 & 0.923 & 0.813 & 0.810 & 0.988 & 0.692 & 0.693 & 0.976 & 0.554 & 0.535 & 0.900 & 0.472 & 0.440 & 0.898 & 0.816 & 0.817 & 0.990 & 0.697 & 0.686 & 0.940 & 0.124 & 0.067 & 0.542 & 0.431 & 0.398 & 0.880 & 0.598 & 0.584 & 0.928 & 0.527 & 0.526 & 0.926 & 0.543 & 0.532 & 0.911 \\
 & LSST & 0.573 & 0.328 & 0.850 & 0.591 & 0.323 & 0.875 & 0.606 & 0.337 & 0.857 & 0.303 & 0.055 & 0.547 & 0.475 & 0.191 & 0.779 & 0.590 & 0.324 & 0.862 & 0.416 & 0.144 & 0.746 & 0.317 & 0.042 & 0.522 & 0.242 & 0.111 & 0.638 & 0.285 & 0.138 & 0.630 & 0.535 & 0.317 & 0.835 & 0.525 & 0.286 & 0.821 \\
 & FingerMovements & 0.494 & 0.447 & 0.511 & 0.492 & 0.454 & 0.474 & 0.510 & 0.416 & 0.531 & 0.492 & 0.482 & 0.517 & 0.538 & 0.509 & 0.555 & 0.502 & 0.437 & 0.508 & 0.502 & 0.487 & 0.506 & 0.500 & 0.371 & 0.502 & 0.472 & 0.445 & 0.468 & 0.544 & 0.544 & 0.588 & 0.520 & 0.519 & 0.531 & 0.556 & 0.553 & 0.563 \\
 & StandWalkJump & 0.360 & 0.269 & 0.619 & 0.387 & 0.252 & 0.572 & 0.280 & 0.167 & 0.468 & 0.533 & 0.537 & 0.673 & 0.413 & 0.387 & 0.572 & 0.280 & 0.168 & 0.447 & 0.440 & 0.395 & 0.569 & 0.333 & 0.167 & 0.500 & 0.333 & 0.225 & 0.499 & 0.360 & 0.352 & 0.603 & 0.533 & 0.508 & 0.805 & 0.560 & 0.546 & 0.771 \\
\bottomrule
\end{tabular}%
}
\end{table*}

\FloatBarrier

\end{document}